%% file: main.tex
\documentclass[10pt,journal,cspaper,compsoc]{IEEEtran}

\usepackage{epsfig}
\usepackage{graphicx} %
\usepackage{amsmath}
\usepackage{amssymb}
\usepackage{array}
\usepackage{enumitem}
\usepackage{multirow}
\usepackage[footnotesize]{subfigure}
\usepackage{xcolor}
\usepackage[normalem]{ulem} %

\usepackage[lined,linesnumbered,ruled]{algorithm2e}
\usepackage{ragged2e} %

\SetCommentSty{mycommfont}
\DeclareMathOperator*{\argmax}{arg\,max}

\DeclareMathOperator*{\softmax}{softmax}

\newcommand{\eg}{\textit{e}.\textit{g}.}
\newcommand{\ie}{\textit{i}.\textit{e}.}
\newcommand{\etal}{\textit{et} \textit{al}.}

\newcolumntype{M}[1]{>{\centering\arraybackslash}m{#1}}
\ifCLASSOPTIONcompsoc
  \usepackage[nocompress]{cite}
\else
  \usepackage{cite}
\fi

\begin{document}
\title{Variational Context: Exploiting Visual and Textual Context for Grounding Referring Expressions}
\author{Yulei Niu, Hanwang Zhang, Zhiwu Lu, Shih-Fu Chang
\thanks{Y. Niu and Z. Lu are with the Beijing Key Laboratory of Big Data Management and Analysis Methods, School of Information, Renmin University of China, Beijing 100872, China. Email: niu@ruc.edu.cn, zhiwu.lu@gmail.com.}
\thanks{H. Zhang is with the School of Computer Science and Engineering,
Nanyang Technological University, Singapore 639798. E-mail:
hanwangzhang@gmail.com.}
\thanks{S.-F. Chang is with the Department of Electrical Engineering, Columbia University, New York, NY 10027, USA. Email: sfchang@ee.columbia.edu.}
}

\hyphenpenalty=10000
\exhyphenpenalty=10000

\IEEEtitleabstractindextext{%
\justify
\begin{abstract}
We focus on grounding (i.e., localizing or linking) referring expressions in images, e.g., ``largest elephant standing behind baby elephant''.  This is a general yet challenging vision-language task since it does not only require the localization of objects, but also the multimodal comprehension of context --- visual attributes (e.g., ``largest'', ``baby'') and relationships (e.g., ``behind'') that help to distinguish the referent from other objects, especially those of the same category. Due to the exponential complexity involved in modeling the context associated with multiple image regions, existing work oversimplifies this task to pairwise region modeling by multiple instance learning. In this paper, we propose a variational Bayesian method, called Variational Context, to solve the problem of complex context modeling in referring expression grounding. Specifically, our framework exploits the \textit{reciprocal relation} between the referent and context, \ie, either of them influences estimation of the posterior distribution of the other, and thereby the search space of context can be greatly reduced. In addition to reciprocity, our framework considers the \textit{semantic information} of context, \ie, the referring expression can be reproduced based on the estimated context. We also extend the model to unsupervised setting where no annotation for the referent is available. Extensive experiments on various benchmarks show consistent improvement over state-of-the-art methods in both supervised and unsupervised settings.
\end{abstract}

\begin{IEEEkeywords}
Grounding referring expression, variational Bayesian model, referring expression generation
\end{IEEEkeywords}
}

\maketitle
\begin{figure}
	\centering
	\includegraphics[width=.9\linewidth]{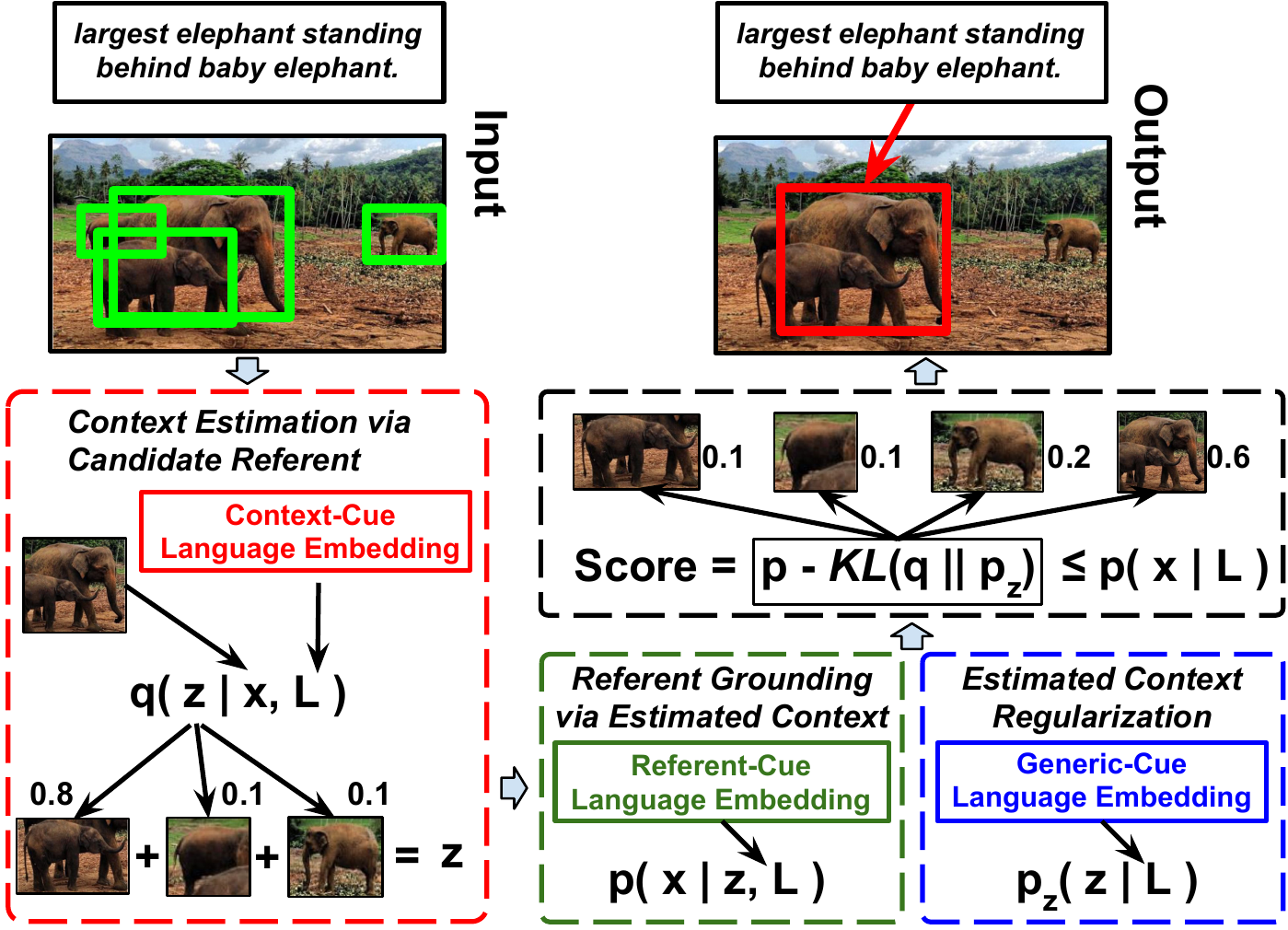}
	\caption{The proposed Variational Context framework. Given an input referring expression and an image with region proposals, we localize the referent as output. We develop a grounding score function, with the variational lower-bound composed by three cue-specific multimodal modules, indicated by the description in the dashed color boxes.}
\vspace{-4mm}
\label{fig:1}
\end{figure}

\IEEEdisplaynontitleabstractindextext
\IEEEpeerreviewmaketitle
\IEEEraisesectionheading{\section{Introduction}\label{sec:introduction}}
\IEEEPARstart{G}rounding natural language in visual data is a hallmark of artificial intelligence, since it establishes a communication channel between humans, machines, and the physical world, underpinning a variety of multimodal artificial intelligence tasks such as robotic navigation~\cite{thomason2017guiding}, visual question answering~\cite{antol2015vqa,li2017visual,zhao2017video}, and visual chatbot~\cite{visdial_rl}. Thanks to the rapid development in deep learning-based computer vision and natural language processing, we have witnessed promising results not only in grounding nouns (\eg, object detection~\cite{redmon2016yolo9000}), but also short phrases (\eg, noun phrases~\cite{plummer2016phrase} and relations~\cite{zhang2016vtranse,sun2017domain}). However, the more general task: grounding referring expressions~\cite{mao2016generation}, is still far from resolved due to the challenges in understanding of both language and scene compositions~\cite{golland2010game}. As illustrated in Fig.~\ref{fig:1}, given an input referring expression ``largest elephant standing behind baby elephant'' and an image with region proposals, a model that can only localize ``elephant'' is not satisfactory as there are multiple elephants. Therefore, the key for referring expression grounding is to comprehend the context. Here, we refer to \emph{context} as the visual objects (\eg, ``elephant''), attributes (\eg, ``largest'' and ``baby''), and relationships (\eg, ``behind'') mentioned in the expression that help to distinguish the referent from other objects.

One straightforward way of modeling the relations between the referent and context is to: 1) use external syntactic parsers to parse the expression into entities, modifiers, and relations~\cite{schuster2015generating}, and then 2) apply visual relation detectors to localize them~\cite{zhang2016vtranse}. However, this two-stage approach is not practical due to the limited generalization ability of the detectors applied in the highly unrestricted language and scene compositions. To this end, recent approaches use multimodal embedding networks that jointly comprehend language and model the visual relations~\cite{nagaraja2016modeling,hu2016modeling}. Due to the prohibitively high cost of annotating both referent and context of referring expressions in images, multiple instance learning (MIL)~\cite{dietterich1997solving} is usually adopted in them to handle the weak supervision of the unannotated context objects, by maximizing the joint likelihood of every region pair. However, for a referent, the MIL framework essentially oversimplifies the number of context configurations of $N$ regions from $\mathcal{O}(2^N)$ to $\mathcal{O}(N)$. For example, to localize the ``elephant'' in Fig.~\ref{fig:1}, we may need to consider the other three elephants all together as a multinomial subset for modeling the context such as ``largest'', ``behind'' and ``baby elephant''. 

In this paper, we propose a novel framework called \emph{Variational Context} for grounding referring expressions in images. Compared to the previous MIL-based approaches~\cite{nagaraja2016modeling,hu2016modeling}, our model approximates the combinatorial context configurations with weak supervision using a variational Bayesian framework~\cite{kingma2013auto}. Intuitively, it exploits the \textit{reciprocity} between referent and context, given either of which can help to localize the other. As shown in Fig.~\ref{fig:1}, for each region $x$, we first estimate a coarse context $z$, which will help to refine the true localizations of the referent. This reciprocity is formulated into the variational lower-bound of the grounding likelihood $p(x|L)$, where $L$ is the text expression and the context is considered as a hidden variable $z$ (cf. Section~\ref{sec:3}). Specifically, the model consists of three multimodal modules: context posterior $q(z|x,L)$, referent posterior $p(x|z, L)$, and context prior $p_z(z|L)$, each of which performs a grounding task (cf. Section~\ref{sec:4_3}) that aligns image regions with a cue-specific language feature; each cue dynamically encodes different subsets of words in the expression $L$ that help the corresponding localization (cf. Section~\ref{sec:4_2}). In addition to reciprocity, our framework considers the \textit{semantic information} of context, \ie, the referring expression can be reproduced based on the referent and its estimated context. Specifically, the context prior $p_z(z|L)$ is resolved into the likelihood $p(L|z)$ for modeling the context-aware referring expression and the prior $p(z)$ following Bayes' theorem. Our proposed framework unifies both referring expression comprehension and generation to promote the context modeling.

Thanks to the reciprocity between referent and context, our model can not only be used in the conventional supervised setting, where there is annotation for referent, but also in the challenging unsupervised setting, where there is no instance-level annotation (\eg, bounding boxes) of both referent and context. We perform extensive experiments on four benchmark referring expression datasets: RefCLEF~\cite{kazemzadeh2014referitgame}, RefCOCO~\cite{yu2016modeling}, RefCOCO+~\cite{yu2016modeling}, and RefCOCOg~\cite{mao2016generation}. Our model consistently outperforms previous methods in both supervised and unsupervised settings. We also qualitatively show that our model can ground the context in the expression to the corresponding image regions (cf. Section~\ref{sec:5}). An earlier version of this work has appeared in~\cite{zhang2018grounding}. The current paper 1) unifies referring expression comprehension and generation to promote complex context modeling, 2) updates state-of-the-art grounding results, 3) enriches qualitative results and failure cases studies, and 4) performs automatic evaluation and human evaluation on referring expression generation.

\section{Related Work}\label{sec:2}
\textbf{Grounding Referring Expression}. Referring expression is the natural language description of a given object or region in an image. Grounding referring expression, also known as referring expression comprehension, intends to localize the target object given the referring expression. Different from grounding phrases~\cite{plummer2015flickr30k,plummer2016phrase} and descriptive sentences~\cite{hu2016natural,rohrbach2016grounding}, the key for grounding referring expression is to use the context (or pragmatics in linguistics~\cite{thomas2014meaning}) to distinguish the referent from other objects, usually of the same category~\cite{golland2010game}. However, most previous works resort to use holistic context such as the entire image~\cite{mao2016generation, hu2016natural,rohrbach2016grounding,deng2018visual} or visual feature difference between regions~\cite{yu2016modeling,yu2016joint,yu2018mattnet}. Our model is similar to the works on explicitly modeling the referent and context region pairs~\cite{hu2016modeling,nagaraja2016modeling}. However, due to the lack of context annotation, they reduce the grounding task into a multiple instance learning framework~\cite{dietterich1997solving}. As we will discuss later, this framework is not a proper approximation to the original task. There are also studies on visual relation detection that detect objects and their relationships~\cite{lu2016visual,Dai_2017_CVPR,zhang2016vtranse,li2017vip,zhang2017ppr,yu2018mattnet}. However, they are limited to a fixed-vocabulary set of relation triplets and hence are difficult to be applied in natural language grounding. Our cue-specific language feature is similar to the language modular network~\cite{hu2016modeling} that learns to decompose a sentence into referent/context-related words, which are different from other approaches that treat the expression as a whole~\cite{mao2016generation,luo2017comprehension,yu2016joint,liureferring}.

\noindent \textbf{Referring Expression Generation}. As the inverse task of grounding referring expression, referring expression generation~\cite{mao2016generation} aims to produce a distinct expression about one object in an image. Different from image captioning~\cite{vinyals2015show,xu2015show}, referring expression generation mainly focuses on one specific object instead of the whole image. In addition, the generated referring expression should be unambiguous and include discriminative information attributes, location and relation. Most of early works have studied referring expression generation in neutral language processing~\cite{krahmer2012computational,mitchell2013generating,winograd1972understanding,van2006building,mitchell2010natural}. However, they focused on the small and artificial datasets of past and have less concern about complex real-world vision problem. Recently, Kazemzadeh \etal~\cite{kazemzadeh2014referitgame} collected a large-scale dataset RefCLEF for natural pictures in real world via a two-player game, where one player provides a referring expression given the object in an image, and another player localizes the referent based on the referring expression and image. Other datasets RefCOCO, RefCOCO+ and RefCOCOg~\cite{yu2016modeling,mao2016generation} on MSCOCO were also collected in the same way. Based on the large-scale datasets, recent works make contributions to linking vision and language. The CNN-RNN structure, widely applied in image captioning, is generally used in referring expression generation. Recent works have investigated the combination of referring expression comprehension and generation. Our proposed Variational Context framework has the following differences in the combination strategy. 1) Compared to~\cite{mao2016generation} that formulated expression generation as $\argmax_L p(L|x,I)$ (where $L$ represents the description, $x$ represents the referent, and $I$ represents the image), we formulate the generation problem as $\argmax_L p(L|x,z(x),I)$, by considering the context $z(x)$ of referent $x$. 2) Compared to~\cite{yu2016modeling} that proposed to jointly generate expressions of objects in the same class for unambiguous expression generation, we separately generate each expression of a referent with its context. 3) Compared to~\cite{yu2016joint} that first accumulated comprehension and generation losses and then formulated the overall loss function as a multi-task learning problem, our proposed framework first formulates visual grounding as a marginal distribution approximation problem, and then resolves the context prior $p(z|L)$ to include generation likelihood $p(L|z)$ in the lower-bound approximation.

\noindent \textbf{Variational Bayesian Model vs. Multiple Instance Learning}.
Our proposed variational context framework is in a similar vein of the deep neural network based variational autoencoder (VAE)~\cite{kingma2013auto}, which uses neural networks to approximate the posterior distribution of the hidden value $q(z|x)$, \ie, encoder, and the conditional distribution of the observation $p(x|z)$, \ie, decoder. VAE shows efficient and effective end-to-end optimization for the \emph{intractable} log-sum likelihood $\log\sum_z p(x,z)$ that is widely used in generative processes such as image synthesis~\cite{yan2016attribute2image} and video frame prediction~\cite{xue2016visual}. Considering the unannotated context as the hidden variable $z$, the referring expression grounding task can also be formulated into the above log-sum marginalization (cf. Eq.~\eqref{eq:2}). The MIL framework~\cite{dietterich1997solving} is essentially a sum-log approximation of the log-sum, \ie, $\sum_z\log p(x,z)$. To see this, the max-pooling function $\log\max_z p(x,z)$ used in~\cite{hu2016modeling} can be viewed as the sum-log $\sum_z\log p(x|z)p(z)$, where $p(z) = 1$ if $z$ is the correct context and 0 otherwise, indicating there is only one positive instance; maximizing the noisy-or function $\log(1-\prod_z(1-p(x,z)))$ used in~\cite{nagaraja2016modeling} is equivalent to maximize $\sum_{z}\log p(x,z)$, assuming there is at least one positive instance. However, due to the numerical property of the log function, this sum-log approximation will unnecessarily force every $(x,z)$ pair to explain the data~\cite{fox2012tutorial}. Instead, we use the variational Bayesian upper-bound to obtain a better sum-log approximation. Note that visual attention models~\cite{ba2014multiple,xu2015show} simplify the variational lower bound by assuming $p(z) = q(z|x)$; however, we explicitly use the KL divergence $K\!L(q(z|x)||p(z))$ in the lower bound to regularize the approximate posterior $q(z|x)$ not being too far from the prior $p(z)$.

\section{Variational Context}\label{sec:3}
In this section, we derive the variational Bayesian formulation of the proposed Variational Context framework and the objective function for training and test.
\subsection{Problem Formulation}
In this paper, we follow the classical definition of grounding referring expressions, where the region proposal generation stage is assumed to be done and only the region retrieval stage is considered. Note that one-stage visual grounding has the potential to be efficient~\cite{deng2019you,chen2018real}, which is out of the discussion in this paper. The classical task of grounding a referring expression $L$ in an image $I$ aims to retrieval the target object $x^*$ from a candidate set of regions $\mathcal{X}$. Formally, we maximize the log-likelihood of the conditional distribution to localize the referent region $x^*\in\mathcal{X}$:
\begin{equation}\label{eq:1}
x^* = \argmax_{x\in\mathcal{X}}\log p(x|L),
\end{equation}
where we omit the image $I$ in $p(x|I, L)$. 

As there is usually no annotation for the context, we consider it as a hidden variable $z$. Therefore, Eq.~\eqref{eq:1} can be rewritten as the following maximization of the log-likelihood of the conditional marginal distribution:
\begin{equation}\label{eq:2}
x^* = \argmax_{x\in\mathcal{X}}\log \sum\limits_{z}p(x,z|L).
\end{equation}
Note that $z$ is NOT necessary to be one region as assumed in recent MIL approaches~\cite{hu2016modeling,nagaraja2016modeling}, \ie, $z\in\mathcal{X}$. For example, the contextual objects ``surrounding elephants'' in ``a bigger elephant than the surrounding elephants'' should be composed by a multinomial subset of $\mathcal{X}$, resulting in an extremely large sample space that requires $\mathcal{O}(2^{|\mathcal{X}|})$ search complexity. Therefore, the marginalization in Eq.~\eqref{eq:2} is intractable in general. 

To this end, we use the variational lower-bound~\cite{kingma2013auto} to approximate the marginal distribution in Eq.~\eqref{eq:2} as:
\begin{equation}\label{eq:3}
\begin{split}
&\log p(x|L) = \log \sum\limits_z p(x,z|L) \geq \mathcal{Q}(x,L)=\\
&\underbrace{\mathbb{E}_{z\sim q_{\phi}(z|x,L)}\log p_{\theta}(x|z,L)}_{\text{Localization }}-\underbrace{K\!L\!\left(q_\phi(z|x,L)||p_\omega(z|L)\right)}_{\text{Regularization}},
\end{split}
\end{equation}
where $K\!L(\cdot)$ is the Kullback-Leibler divergence, $\phi$, $\theta$, and $\omega$ are independent parameter sets for the respective distributions. As shown in Fig.~\ref{fig:1}, the lower bound $\mathcal{Q}(x, L)$ offers a new perspective for exploiting the reciprocal nature of referent and context in referring expression grounding.

\subsubsection{Localization} This term calculates the localization score for $x$ given an estimated context $z$, using the referent-cue of $L$ parameterized by $\theta$. In particular, we design a new posterior $q_\phi(z|x,L)$ that approximates the true context posterior $p(z|x, L)$, which models the context $z$ using the context-cue of $L$ parameterized by $\phi$. In the view of variational auto-encoder~\cite{kingma2013auto,sohn2015learning}, this term works in an encoding-decoding fashion: $q_\phi$ is the encoder from $x$ to $z$, and $p_\theta$ is the decoder from $z$ to $x$. 

\subsubsection{Regularization} Since the Kullback-Leibler divergence ($K\!L$) is non-negative, maximizing $\mathcal{Q}(x,L)$ would encourage that the posterior $q_\phi$ is similar to the prior $p_\omega$, \ie, \textit{the estimated context $z$ sampled from $q_\phi (z|x, L)$ should not be too far from the referring expression}, which is modeled by $p_\omega(z|L)$ with the generic-cue of $L$ parameterized by $\omega$. This term is necessary as the estimated context $z$ could be overfitted to region features that are inconsistent with the visual context described in the expression.

Furthermore, we hope that the estimated context $z$ contains necessary semantic information, which can helps to reproduce the referring expression. This can be realized by unifying both referring expression comprehension and generation, and resolving the context prior $p(z|L)$ as:
\begin{equation}\label{eq:pzl}
    p(z|L)=g(x,L)p(L|z),
\end{equation}
where $g(x,L)$ is the function representing $p(z)/p(L)$, and we omit the referent region $x$ in $z(x)$ for simplicity. The likelihood $p(L|z)$ models the referring expression $L$ based on the estimated context $z$. Applying Eq.~\eqref{eq:pzl} to Eq.~\eqref{eq:3}, we can get another lower bound $\mathcal{Q}'(x,L)$:
\begin{equation}
\begin{split}
    \mathcal{Q}'(x,L)
    =&\mathbb{E}_{z\sim q_{\phi}(z|x,L)}[\log p_{\theta}(x|z,L)-\log q_\phi(z|x,L)\!\\
    &+\log g(x,L)+\log p(L|z)],
\end{split}
\end{equation}

\begin{figure*}
	\centering
	\includegraphics[width=1\linewidth]{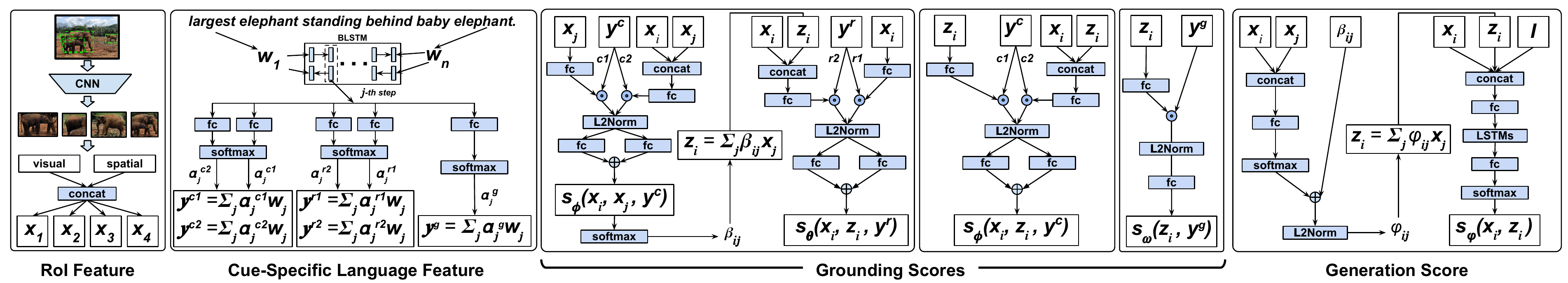}
	\caption{The architecture of the proposed Variational Context framework. It consists of a region feature extraction module (Section~\ref{sec:4_1}), and a language feature extraction module (Section~\ref{sec:4_2}), three grounding modules (Section~\ref{sec:4_3}), and one generation module (Section~\ref{sec:4_4}). It can be trained in an end-to-end fashion with the input of a set of image regions and a referring expression, using the supervised loss (Eq.~\eqref{eq:7}) or the unsupervised loss (Eq.~\eqref{eq:8}). LSTMs: Long short-term memory networks. fc: fully-connected layer. concat: vector concatenation. L2Norm: L2 normalization layer. $\odot$: element-wise vector multiplication. $\oplus$: add.}
\vspace{-1mm}
\label{fig:2}
\end{figure*}

\subsection{Training and Test}

\noindent \textbf{Deterministic Context}. The lower-bound $\mathcal{Q}(x, L)$ transforms the intractable log-sum in Eq.~\eqref{eq:2} into the efficient sum-log in Eq.~\eqref{eq:3}, which can be optimized by using Monte Carlo unbiased gradient estimator such as REINFORCE~\cite{williams1992simple}. However, due to that $\phi$ is dependent on the sampling of $z$ over $\mathcal{O}(2^{|\mathcal{X}|})$ configurations, its gradient variance is large. To this end, we implement $q_\phi(z|x,L)$ as a differentiable but biased encoder:
\begin{equation}\label{eq:4} 
z = f(x,L) = \sum\limits_{x'\in\mathcal{X}}x'\cdot q_\phi(x'|x,L), 
\end{equation}
where we slightly abuse $q_\phi$ as a score function such that $\sum_{x'}q_\phi(x'|x,L) = 1$. Note that this deterministic context can be viewed as applying the ``re-parameterization'' trick as in Variational Auto-Encoder~\cite{kingma2013auto}: rewriting $z\sim q_\phi(z|x,L)$ to $z = f(x,L;\epsilon), \epsilon\sim p(\epsilon)$, where the stochasticity of the auxiliary random variable $\epsilon$ comes from training samples $x\in\mathcal{X}(\epsilon)$.  A clear example is Adversarial Autoencoder~\cite{aae} which shows that such stochasticity achieves similar test-likelihood compared to other distributions. 

\noindent \textbf{Objective Function}. Applying Eq.~\eqref{eq:4} to Eq.~\eqref{eq:3}, we can rewrite $Q(x,L)$ into a function of only one sample estimation, which is a common practice in SGD:
\begin{equation}\label{eq:5}
\begin{split}
\!\mathcal{Q}(x,\! L) &\!=\! \log p_\theta(x|z,\! L)\!-\!\log q_\phi(z|x,\! L)\!+\!\log p_\omega(z|L).
\end{split}
\end{equation}
In supervised setting where the ground truth of the referent is known, to distinguish the referent from other objects, we need to train a model that outputs a high $p(x|L)$ (\ie, $\mathcal{Q}(x,L)$), while maintaining a low $p(x'|L)$ (\ie, $\mathcal{Q}(x',L)$), whenever $x'\!\neq\!x$. Therefore, we use the so-called Maximum Mutual Information loss as in~\cite{mao2016generation} $-\log\{\mathcal{Q}(x,L)/\sum_{x'}\mathcal{Q}(x',L)\}$, where we do not need to explicitly model the distributions with normalizations; we use the following score function:
\begin{equation}\label{eq:6}
\mathcal{Q}(x,L)\propto\mathcal{S}(x,L) = s_\theta(x, L)-s_\phi(x, L)+s_\omega(x, L),
\end{equation}
where $z$ is omitted as it is a function of $x$ in Eq.~\eqref{eq:4}. $s_\theta$, $s_\phi$, and $s_\omega$ are the score functions (\eg, $p_\theta\!\propto\!s_\theta$) for $p_\theta$, $q_\phi$, and $p_\omega$, respectively. These functions will be detailed in Section~\ref{sec:4_3}. Similar to Eq.~\eqref{eq:6}, we use the following score function to incorporate referring expression generation in the variational framework:
\begin{equation}\label{eq:5-2}
\begin{split}
\mathcal{Q}'\!(x,L)\propto\mathcal{S}'\!(x,L)&=s_\theta(x,L)-s_\phi(x,L)+s_{\omega'}(x,L)\!\\
&\quad+s_\psi(x,L),
\end{split}
\end{equation}
where $s_{\omega'}$ is the score function for $g(x,L)$, and shares the same structure with $s_\omega$; $s_{\psi}$ is the score function for $p(L|z)$.

In this way, maximizing Eq.~\eqref{eq:5} is equivalent to minimizing the following softmax loss:
\begin{equation}\label{eq:7}
\mathcal{L}_{s} = -\log\softmax \mathcal{S}(x_{gt},L),
\end{equation}
where the softmax is over $x\in\mathcal{X}$ and $x_{gt}$ is the ground truth referent region.

Note that the reciprocity between referent and context can be extended to unsupervised learning, where neither of the referent and context has annotation. In this setting, we adopt the \emph{image-level} max-pooled MIL loss functions for unsupervised referring expression grounding:
\begin{equation}\label{eq:8}
\mathcal{L}_{u} = -\max\limits_{x\in\mathcal{X}}\log\softmax\mathcal{S}(x,L),
\end{equation}
where the softmax is over $x\in\mathcal{X}$. Note that the max-pooled MIL function is reasonable since there is only one ground truth referent given an expression and image training pair. 

At test stage, in both supervised and unsupervised settings, we predict the referent region $x^*$ by selecting the region $x\in\mathcal{X}$ with the highest score:
\begin{equation}\label{eq:9}
x^* = \argmax\limits_{x\in\mathcal{X}}\mathcal{S}(x,L).
\end{equation}

\noindent \textbf{Referring Expression Generation}. During training stage, instead of using the ground-truth referent, we sample a concrete referent region $\hat{x}$ from $p(x|L)$ and calculate its estimated context $\hat{z}\!=\!f(\hat{x},L)$ using Eq.~\eqref{eq:4}. The reason why we use the sampled referent is to punish false grounding results using the generation module. However, the generation loss $\mathcal{L}_G\!=\mathbb{E}_{x\sim p(x|L)}\mathcal{L}_{c}(x,L)$ is non-differentiable to referent grounding part using the concrete referent, where $\mathcal{L}_{c}(x,L)$ is the sum of cross entropy over the predicted words at each step. Hence, we apply policy gradient method in REINFORCE~\cite{williams1992simple} for end-to-end training. The gradient $\nabla\mathcal{L}_G$ of the generation loss $\mathcal{L}_G$ is:
\begin{equation}\label{eq:rl}
\nabla\mathcal{L}_G\!=\!E_{x\sim p(x|L)}[\mathcal{L}_{c}(x,L)\nabla\log p(x|L)+\nabla\mathcal{L}_{c}(x,L)].
\end{equation}
In practice, the gradient $\nabla\mathcal{L}_G$ can be estimated using Monte-Carlo sampling as:
\begin{equation}\label{eq:rl}
\nabla\!\mathcal{L}_G\!\approx\! \frac{1}{K}\!\sum^N_{k=1}[\mathcal{L}_{c}\!\left(x_k,L\right)\!\nabla\!\log p\!\left(x_k|L\right)\!+\!\nabla\!\mathcal{L}_{c}\!\left(x_k,L\right)],
\end{equation}
where $x_{k}$ is sampled from $p(x|L)$. We simply use $K\!=\!1$ in our implementation. Since $p(x|L)$ is fully differentiable, the gradient can be transferred to referent grounding part via backpropagation. Following \cite{weaver2001optimal}, we utilize a moving average baseline $b$ to reduce the variance of estimated gradient using REINFORCE, and replace $\mathcal{L}_{c}(x_k,L)$ with $\mathcal{L}_{c}(x_k,L)\!-\!b$ in Eq. (\ref{eq:rl}). The baseline $b_t$ at $t$-th iteration is estimated by accumulating the previous losses $\mathcal{L}_c(x,L)$ with exponential decay:
\begin{equation}
b_t = 0.9\times b_{t-1}+0.1\times \mathcal{L}_{c}(x_{k_t},L).
\end{equation}

\section{Model Architecture}
The overall architecture of the proposed variational context framework is illustrated in Fig.~\ref{fig:2}.  Thanks to the deterministic context in Eq.~\eqref{eq:4} and REINFORCE in Eq.~\eqref{eq:rl}, the six modules in our model can be integrated into an end-to-end differentiable fashion. Next, we will detail the implementation of each module.  

\subsection{RoI Features}\label{sec:4_1}
Given an image with a set of Region of Interests (RoIs) $\mathcal{X}$, obtained by any off-the-shelf proposal generator~\cite{zitnick2014edge} or object detectors~\cite{liu2016ssd}, this module extracts the feature vector $\mathbf{x}_i$ for every RoI. In particular, $\mathbf{x}_i$ is the concatenation of visual feature $\mathbf{v}_i$ and spatial feature $\mathbf{p}_i$. For $\mathbf{v}_i$, we can use the output of a pre-trained convolutional network (cf. Section~\ref{sec:5}). If the object category of each RoI is available, we can further utilize the comparison between the referent and other objects to capture the visual difference such as ``the largest/baby elephant''. Specifically, we append the visual difference (visdif) feature~\cite{yu2016modeling} $\delta \mathbf{v}_i = \frac{1}{n}\sum_{j\neq i}\frac{\mathbf{v}_i-\mathbf{v}_j}{||\mathbf{v}_i-\mathbf{v}_j||}$ to the original $\mathbf{v}_i$ visual feature, where $n$ is the number of objects chosen for comparison (\eg, the number of RoI in the same object category). For spatial feature, we use the 5-d spatial attributes $\mathbf{p}_i = [\frac{x_{tl}}{W}, \frac{y_{tl}}{H}, \frac{x_{br}}{W}, \frac{y_{br}}{H}, \frac{w\cdot h}{W\cdot H}]$, where $x$ and $y$ are the coordinates the top left (tl) and bottom right (br) RoI of the size $w\times h$, and the image is of the size $W\times H$. 

\begin{figure}
	\centering
	\includegraphics[width=0.98\linewidth]{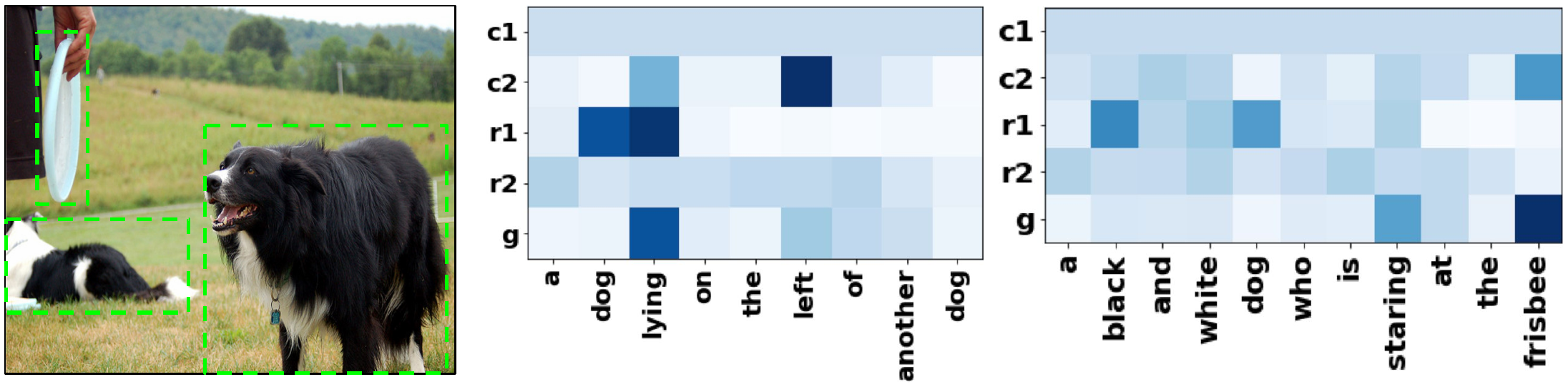}
	\caption{Two qualitative examples of the cue-specific language feature word weights. Darker color indicates higher weights. c/r+1/2: context/referent-cue + single/pairwise. }
\label{fig:3}
\vspace{-1mm}
\end{figure}

\subsection{Cue-Specific Language Features}\label{sec:4_2}
An alternative for language understanding is to employ external NLP parsers to parse the referring expression into several textual components, such as subject, object, and relationship. However, conventional parsers (\eg, Standford Dependency) are observed to be suboptimal to the grounding referring expression task~\cite{hu2016modeling}, and cannot be end-to-end trained. In this work, we choose to parse the referring expression with language attention by learning in an end-to-end fashion. Specifically, the referring expression is represented into different cue-specific language features, which is inspired by the attention weighted sum of word vectors~\cite{hu2016modeling, lu2016hierarchical,bahdanau2014neural}. We parameter the weights by context-cue $\phi$, referent-cue $\theta$, and generic-cue $\omega$ based on their different purposes: the context-cue feature helps to estimate context prior, the referent-cue feature aims to localize the referent, and the generic-cue feature encourages the estimated context to be consistent with the visual context described in the expression. Formally, the context-cue language feature $\mathbf{y}^c = [\mathbf{y}^{c1}, \mathbf{y}^{c2}]$ is a concatenation of $\mathbf{y}^{c1}$ for language-vision association between \emph{single} RoI and the expression, and $\mathbf{y}^{c2}$ for the association between \emph{pairwise} RoIs; the referent-cue language feature $\mathbf{y}^r$ can be represented in a similar way to $\mathbf{y}^c$; the generic-cue language feature $\mathbf{y}^g$ is only for single RoI association as it is an independent prior. The weights of each cue are calculated from the hidden state vectors of a 2-layer bi-directional LSTM (BLSTM)~\cite{schuster1997bidirectional}, scanning through the expression. The hidden states encode forward and backward compositional semantic meanings of the sentences, beneficial for selecting words that are useful for single and pairwise associations. Specifically, suppose $\mathbf{h}_j$ as the 4,000-d concatenation of forward and backward hidden vectors of the $j$-th word, without loss of generality, the word attention weight $\alpha_j$ and the language feature $\mathbf{y}$ for single/pairwise association of any cue can be calculated as:
\begin{equation}\label{eq:10}
\mathbf{m}_j = \textrm{fc}(\mathbf{h}_j), \alpha_{j} = \textrm{softmax}_j(\mathbf{m}_{j}),\mathbf{y} = \sum\nolimits_{j}\alpha_j\mathbf{w}_j,
\end{equation}
where $\mathbf{w}_j$ is a 300-d vector. Note that the BLSTM module can be jointly trained with the entire model. 

Fig.~\ref{fig:3} shows that the cue-specific language features dynamically weight words in different expressions. We can have two interesting observations. First, c1 is almost uniform while c2 is highly skewed; although r2 is more skewed than c1, it is still less skewed than r1. This is reasonable since: 1) without ground-truth, individual score (c1) does not help much for context estimation from scratch; context is more easily found by the pairwise score (c2) induced by relationships or other objects (\eg, ``left'' or ``frisbee''); 2) in referent grounding with ground truth, individual score (r1) is sufficient (\eg, ``dog lying'' and ``black white dog'') and pairwise score (r2) is helpful; 3) g is adaptive to the number of object categories in the expression, \ie, if the context object is of the same category as the referent, g weighs descriptive or relationship words higher (\eg, ``lying, standing, left''), and nouns higher (\eg, ``frisbee''), otherwise; moreover, it demonstrates that the deterministic guess of $z$ in Eq.~\eqref{eq:4} is meaningful.

\subsection{Score Functions for Comprehension}\label{sec:4_3}
For any image and expression pair, given the RoI feature $\mathbf{x}_i$, and the cue-specific language feature $\mathbf{y}^c$, $\mathbf{y}^r$, and $\mathbf{y}^g$, we implement the final grounding score in Eq.~\eqref{eq:6} as:
\begin{equation}\label{eq:11x}
\begin{split}
\mathbf{z}_i = \sum\nolimits_j \textrm{softmax}_j&\left(s_\phi(\mathbf{x}_i,\mathbf{x}_j,\mathbf{y}^c)\right)\mathbf{x}_j,\\
s_\theta(x,L) &\leftarrow s_\theta(\mathbf{x}_i,\mathbf{z}_i,\mathbf{y}^r),\\
s_\phi(x,L) &\leftarrow s_\phi(\mathbf{x}_i,\mathbf{z}_i,\mathbf{y}^c),\\
s_\omega(x,L) &\leftarrow s_\omega(\mathbf{z}_i,\mathbf{y}^g),
\end{split}
\end{equation}
where the right-hand side functions are defined as below.

\textbf{Context Estimation Score}: $s_\phi(\mathbf{x}_i,\mathbf{x}_j,\mathbf{y}^c)$. It is a score function for modeling the context posterior $q_\phi(z|x,L)$, \ie, given an RoI $\mathbf{x}_i$ as the candidate referent, we calculate the likelihood of any RoI $\mathbf{x}_j$ to be the context. We can also use this function to estimate the final context posterior score $s_\phi(\mathbf{x}_i,\mathbf{z}_i,\mathbf{y}^c)$. Specifically, the context estimation score is a sum of the single and pairwise vision-language association scores: $\mathbf{x}_j$ and $\mathbf{y}^{c1}$, $[\mathbf{x}_i,\mathbf{x}_j]$ and $\mathbf{y}^{c2}$. Each associate score is an fc output from the input of a normalized feature:
\begin{equation}\label{eq:11}
\begin{split}
&\mathbf{m}^{1}_{j} = \mathbf{y}^{c1} \odot \textrm{fc}(\mathbf{x}_j),~\mathbf{m}^{2}_{j} = \mathbf{y}^{c2} \odot \textrm{fc}([\mathbf{x}_i,\mathbf{x}_j]),\\
&\widetilde{\mathbf{m}}^1_{j} = \textrm{L2Norm}(\mathbf{m}^{1}_{j}),~\widetilde{\mathbf{m}}^{2}_{j}= \textrm{L2Norm}(\mathbf{m}^{2}_{j}),\\
&s_\phi(\mathbf{x}_i,\mathbf{x}_j,\mathbf{y}^c) = \textrm{fc}(\widetilde{\mathbf{m}}^1_{j})+\textrm{fc}(\widetilde{\mathbf{m}}^{2}_{j}),
\end{split}
 \end{equation}
where the element-wise multiplication $\odot$ is an effective way for fusing multimodal features~\cite{ba2014multiple}. According to Eq.~\eqref{eq:4}, we can obtain the estimated context $z$ as $\mathbf{z}_i = \sum\nolimits_j\beta_{j}\mathbf{x}_j$, where $\beta_{j} = \textrm{softmax}_j(s_\phi(\mathbf{x}_i,\mathbf{x}_j,\mathbf{y}^c))$.

\textbf{Referent Grounding Score}: $s_\theta(\mathbf{x}_i,\mathbf{z}_i,\mathbf{y}^r)$. After obtaining the context feature $\mathbf{z}_i$, we use this score function to calculate how likely a candidate RoI $\mathbf{x}_i$ is the referent given the context $\mathbf{z}_i$. This function is similar to Eq.~\eqref{eq:11}.

\textbf{Context Regularization Score}: $s_\omega(\mathbf{z}_i,\mathbf{y}^g)-s_\phi(\mathbf{x}_i,\mathbf{z}_i,\mathbf{y}^c)$. As discussed in Eq.~\eqref{eq:6}, this function scores how likely the estimated context feature $\mathbf{z}_i$ is consistent with the content mentioned in the expression. In particular, $s_\omega(\mathbf{z}_i,\mathbf{y}^g)$ is only dependent on single RoI:
\vspace{-1mm}
\begin{equation}\label{eq:12}
\mathbf{m}_{i}\!\! =\!\! \mathbf{y}^g_i \odot \textrm{fc}(\mathbf{z}_i),
\widetilde{\mathbf{m}}_{i} \!\!=\!\! \textrm{L2Norm}(\mathbf{m}_i),s_\omega(\mathbf{z}_i,\mathbf{y}^g_i)\!\!=\!\!\textrm{fc}(\mathbf{m}_i).
\end{equation}

\subsection{Score Function for Generation}\label{sec:4_4}
For any referent and expression pair, given the RoI feature $\mathbf{x}_i$ and the context-specific language feature $\mathbf{y}^c$, we implement the generation score function in Eq.~\eqref{eq:5-2} to reconstruct the referring expression as:
\begin{equation}\label{eq:4.4}
\begin{split}
\hat{\mathbf{z}}_i = &\sum\nolimits_j \varphi_{j}\mathbf{x}_j,\\
s_\psi(x,L) &\leftarrow s_\psi(\mathbf{x}_i,\hat{\mathbf{z}}_i),\\
\end{split}
\end{equation}
where $\hat{\mathbf{z}}_i$ represents the estimated context for generation, and the joint region attention $\varphi_{j}$ is defined as below:
\begin{equation}\label{eq:lstm-att}
\begin{split}
\!\!\!\!\beta_{j}\!&=\!\softmax\nolimits_j(s_\phi(\mathbf{x}_i,\mathbf{x}_j,\mathbf{y}^c)),\\
\!\!\!\!\gamma_{j}\!&=\!\softmax\nolimits_j(\textrm{fc}([\mathbf{x}_i, \mathbf{x}_j])),\\
\!\!\!\!\varphi_{j}\!&=\!\textrm{L2Norm}_j(\beta_{j} \odot \gamma_{j}), \\
\end{split}
\end{equation}
where $\beta_{j}$ and $\gamma_{j}$ represents the region attention weight for comprehension and generation respectively, and $\varphi_{j}$ mixes these attention weights using the element-wise multiplication $\odot$. Note that $\beta_j$ shares the same calculation with context estimation score in the comprehension module, which can evaluate the estimated context. We then modify the vanilla language generation model~\cite{vinyals2015show} to reconstruct the referring expression. Different from~\cite{vinyals2015show}, we feed the referent $\mathbf{x}_i$ with its context $\mathbf{\hat{z}}_{i}$ and the image feature $\mathbf{I}$ into the LSTM model for sequence generation:
\begin{equation}\label{eq:lstm}
\begin{split}
&\mathbf{w}_{-\!1}\!=\!\textrm{fc}([\mathbf{x}_i,\mathbf{\hat{z}}_i, \mathbf{I}]),~\mathbf{h}_{-2}\!=\!\mathbf{0},\\
&\mathbf{w}_{t}\!=\!W_e S_t,\mathbf{h}_{t}\!=\!\textrm{LSTM}(\mathbf{w}_t,\mathbf{h}_{t-1}),\\
&\mathbf{p}_{t}\!=\!\softmax(\textrm{fc}(\mathbf{h}_{t})),\\
&s_\psi(\mathbf{x}_i,\hat{\mathbf{z}}_i)=\prod\nolimits_t \mathbf{p}_t^T S_{t+1}
\end{split}
\end{equation}
where $W_e$ is the word embedding matrix shared with the comprehension module, and $S_t$ is the one-hot encoding for the input word $w_t$ at step $t$. Note that the start word and stop word are denoted by $w_0$ and $w_{T\!+\!1}$, respectively, standing for the beginning and end of the referring expression, where $T$ is the length of the referring expression. A complete referring expression is generated when the LSTM encounters the stop word or the length of expression reaches the maximum number.

\section{Experiment}\label{sec:5}
\subsection{Datasets}
We used four popular benchmarks for the referring expression grounding task.

\textbf{RefCOCO}~\cite{yu2016modeling}. It has 142,210 referring expressions for 50,000 referents (\eg, object instances) in 19,994 images from MSCOCO~\cite{lin2014microsoft}. The expressions are collected in an interactive way~\cite{kazemzadeh2014referitgame}. The dataset is split into train, validation, Test A, and Test B, which has 120,624, 10,834, 5,657 and 5,095 expression-referent pairs, respectively. An image contains multiple people in Test A and multiple objects in Test B.

\textbf{RefCOCO+}~\cite{yu2016modeling}. It has 141,564 expressions for 49,856 referents in 19,992 images from MSCOCO. The difference from RefCOCO is that it only allows appearances but no locations to describe the referents. The split is 120,191, 10,758, 5,726 and 4,889 expression-referent pairs for train, validation, Test A, and Test B respectively. 

\textbf{RefCOCOg}~\cite{mao2016generation,yu2018mattnet}. It has 95,010 referring expressions for 49,822 objects in 25,799 images from MSCOCO. Different from RefCOCO and RefCOCO+, this dataset not collected in an interactive way and contains longer sentences containing both appearance and location expressions. RefCOCOg has two types of split. The old split \cite{mao2016generation} is 85,474 and 9,536 expression-referent pairs for training and validation. It should be noticed that the old partitioned by objects, thus some images may exist in both train and validation sets. We represent the validation set as ``Val*''. The new partition \cite{yu2018mattnet} randomly divides images into train, validation and test set. The split is 80,512, 4,896 and 9,602 expression-referent pairs for train, validation and test, respectively. The validation and test sets are represented as ``Val'' and ``Test''. Compared to RefCOCO and RefCOCO+, RefCOCOg contains longer expressions, which makes it more challenging for comprehension and generation.

\textbf{RefCLEF}~\cite{kazemzadeh2014referitgame}. It contains 20,000 images with annotated image regions. It has some ambiguous (e.g. “anywhere”) phrases and mistakenly annotated image regions that are not described in the expressions. For fair comparison, we used the split released by~\cite{hu2016natural,rohrbach2016grounding}, \ie, 58,838, 6,333 and 65,193 expression-referent pairs for training, validation and test, respectively.

\begin{figure*}[t]
	\centering
	\includegraphics[width=0.92\linewidth]{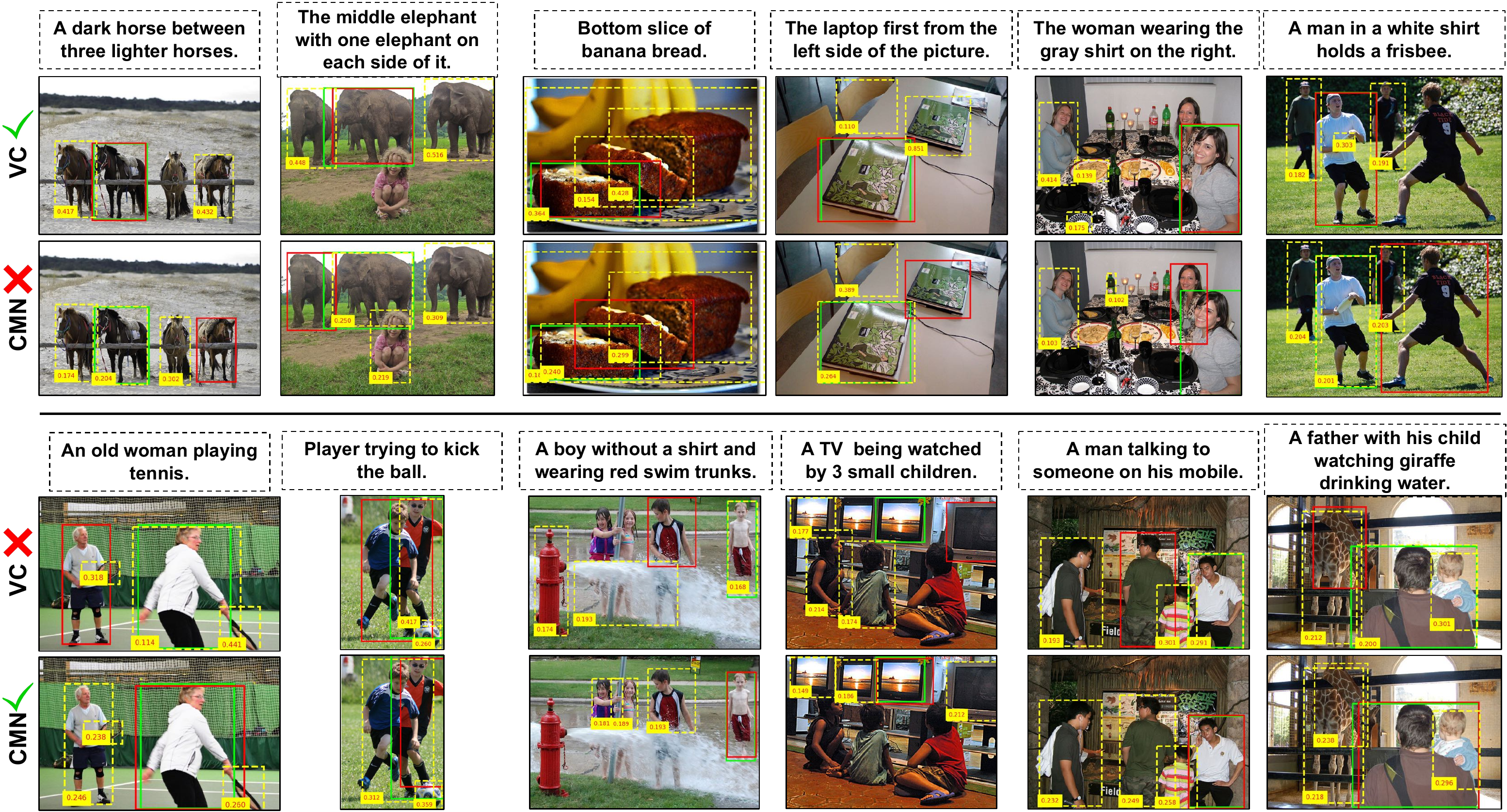}
	\caption{Qualitative results on RefCOCOg (det) showing comparisons between correct (green tick) and wrong referent grounds (red cross) by VC and CMN using VGG features. The denotations of the bounding box colors are as follows. Solid red: grounding referent; solid green: ground truth; dashed yellow: grounding context. We only display top 3 context objects with the context ground probability $>0.1$. We can observe that VC has more reasonable context localizations than CMN, even in cases when the referent ground of VC fails.}
\vspace{-1mm}
\label{fig:4}
\end{figure*}

\begin{figure*}
	\centering
	\includegraphics[width=0.92\linewidth]{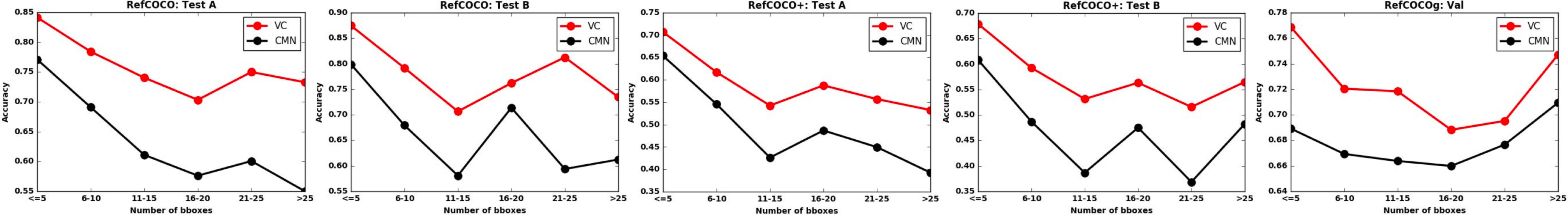}
	\caption{Performances of VC and CMN with different number of object bounding boxes on RefCOCO Test A \&B, RefCOCO+ Test A \& B, and RefCOCOg Val. Compared to CMN, we can see that VC is more effective in context modeling when the number of objects is large.}
\label{fig:5}
\end{figure*}

\subsection{Settings and Metrics}
For comprehension module, we used an English vocabulary of 72,704 words contained in the GloVe pre-trained word vectors~\cite{pennington2014glove}, which was also used for the initialization of our word vectors. We used a ``unk'' symbol for the input word of the BLSTM if the word is out of the vocabulary; we set the sentence length to 20 and used ``pad'' symbol to pad expression sentence $<20$. For RoI visual features on RefCOCO, RefCOCO+, and RefCOCOg which have MSCOCO annotated regions with object categories, we used the concatenation of the 4,096-d fc7 output of a VGG-16 based Faster-RCNN network~\cite{ren2015faster} trained on MSCOCO and its corresponding 4,096-d visdif feature~\cite{yu2016modeling}; although RefCLEF regions also have object categories, for fair comparison with~\cite{rohrbach2016grounding}, we did not use the visdif feature. For generation module, we built an additional vocabulary including words which occur at least 5 times in the training set. The maximum length of generated sentences is set as 20. The hidden state size of LSTM is set as 512. We also regularized the LSTM using dropout with ratio of 0.3. Following \cite{hu2017learning}, we also use an entropy regularization $5\!\times\!{10}^{-3}$ over the conditional distribution $p(x|L)$ to encourage exploration through the sampling space. For fair comparison with \cite{yu2018mattnet}, we also used the average-pooled C4 feature and phrase-guided embedding, which are provided by \cite{yu2018mattnet}.

The model training is single-image based, with all referring expressions annotated. We applied SGD of 0.95-momentum with initial learning rate of 0.01, multiplied by 0.1 after every 120,000 iterations, up to 160,000 iterations. Parameters in BILSTM and fc-layers were initialized by Xavier~\cite{glorot2010understanding} with $5\!\times\!10^{-4}$ weight decay. Other settings were default in TensorFlow. Note that our model is trained without bells and whistles, therefore, other optimization tricks such as batch normalization~\cite{ioffe2015batch} and GRU~\cite{cho2014properties} are expected to further improve the results reported here. Besides the ground truth annotations, grounding to automatically detected objects is a more practical setting. Therefore, we also evaluated with detected objects, the SSD-detected bounding boxes~\cite{liu2016ssd} provided by~\cite{yu2016joint} using VGG-based model, and Faster R-CNN detected bounding boxes provided by \cite{yu2018mattnet} using ResNet-based model. A grounding is considered as correct if the intersection-over-union (IoU) of the top-1 scored region and the ground-truth object is larger than $0.5$. The grounding accuracy (a.k.a, P@1) is the fraction of correctly grounded test expressions.

\begin{figure*}[t]
	\centering
	\includegraphics[width=0.97\linewidth]{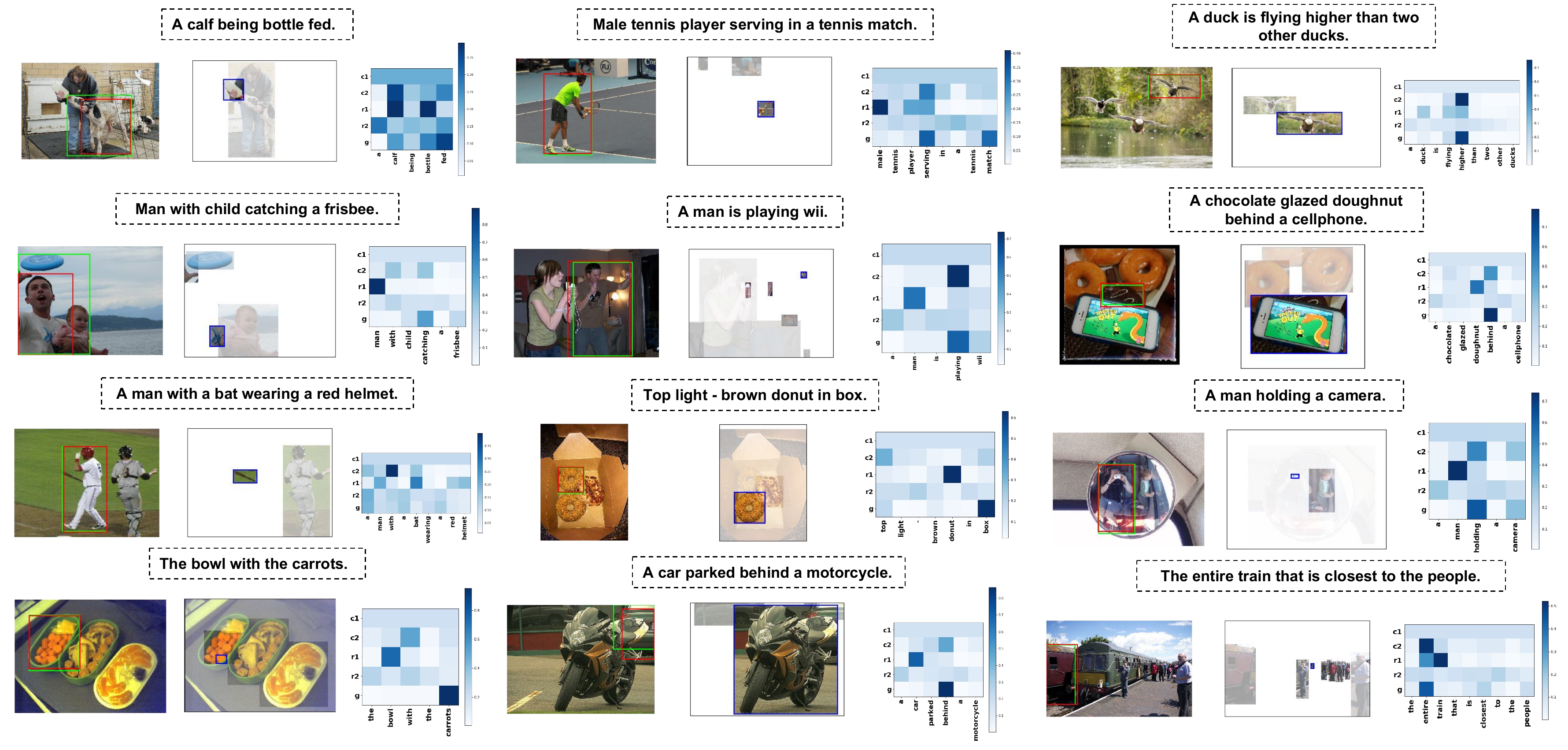}
	\caption{Qualitative results of our full model (VC w/ Gen+PG) on RefCOCOg (det). The first column shows the grounding results. The second column shows the context estimation results. The third column shows the cue-specific language feature word weights. The denotations of the bounding box colors are as follows. Solid red: grounding referent; solid green: ground truth; solid blue: grounding context with highest probability.}
\vspace{-3mm}
\label{fig:6}
\end{figure*}

\begin{table}[t]
\centering
\caption{Supervised grounding performances (Acc\%) of comparing methods using VGG features on MSCOCO ground-truth regions. Note that~\cite{yu2016joint} reports slightly higher accuracies using ensemble models of Listener and Speaker. For fair comparison, we only report their single models.}
\label{tab:1-2}
\scalebox{.92}{
\begin{tabular}{|c|c|c|c|c|c|c|}
\hline 
Dataset & \multicolumn{2}{c|}{RefCOCO} & \multicolumn{2}{c|}{RefCOCO+} & RefCOCOg\\
\hline
Split & Test A & Test B & Test A & Test B & Val* \\
\hline
MMI~\cite{mao2016generation} & 71.72 & 71.09 & 58.42 & 51.23 & 62.14 \\
\hline
NegBag~\cite{nagaraja2016modeling} & 75.6 & 78.0 & --- & --- & 68.4\\
\hline
Attr~\cite{liureferring} & 78.85 & 78.07 & 61.47 & 57.22 & 69.83\\
\hline
CMN~\cite{hu2016modeling} & 75.94 & 79.57 & 59.29 & 59.34 & 69.30\\
\hline
Speaker~\cite{yu2016joint} & 78.95 & 80.22 & 64.60 & 59.62 & 72.63\\
\hline
Listener~\cite{yu2016joint} & 78.45 & 80.10 & 63.34 & 58.91 & 72.25\\
\hline
PLAN~\cite{zhuang2018parallel} & 80.81 & 81.32 & 66.31 & 61.46 & 69.47\\
\hline
A-ATT~\cite{deng2018visual} & \bf 81.17 & 80.01 & \bf 68.76 & 60.63 & 73.18\\
\hline
MAttNet~\cite{yu2018mattnet} & 79.99 & 82.30 & 65.04 & 61.77 & 73.08\\
\hline
\hline
VC w/o reg & 75.59 & 79.69 & 60.76 & 60.14 & 71.05 \\
\hline
VC w/o $\alpha$ & 74.03 & 78.27 & 57.61 & 54.37 & 65.13 \\
\hline
VC & 78.98 & \bf 82.39 & 62.56 & \bf 62.90 & \bf 73.98 \\
\hline
\end{tabular}
}
\vspace{-4mm}
\end{table}

\begin{table}[t]
\centering
\caption{Supervised grounding performances (Acc\%) of comparing methods using VGG features on MSCOCO detected regions. Note that~\cite{yu2016joint} reports slightly higher accuracies using ensemble models of Listener and Speaker. For fair comparison, we only report their single models.}
\label{tab:1-3}
\scalebox{.92}{
\begin{tabular}{|c|c|c|c|c|c|}
\hline 
Dataset & \multicolumn{2}{c|}{RefCOCO} & \multicolumn{2}{c|}{RefCOCO+} & RefCOCOg \\
\hline
Split & Test A & Test B & Test A & Test B & Val* \\
\hline
MMI~\cite{mao2016generation} & 64.90 & 54.51 & 54.03 & 42.81 & 45.85\\
\hline
NegBag~\cite{nagaraja2016modeling} & 58.6 & 56.4 & --- & --- & 39.5\\
\hline
Attr~\cite{liureferring} & 72.08 & 57.29 & 57.97 & 46.20 & 52.35\\
\hline
CMN~\cite{hu2016modeling} & 71.03 & 65.77 & 54.32 & 47.76 & 57.47\\
\hline
Speaker~\cite{yu2016joint} & 72.95 & 63.43 & \bf 60.43 & 48.74 & 59.51\\
\hline
Listener~\cite{yu2016joint} & 72.95 & 62.98 & 59.61 & 48.44 & 58.32\\
\hline
PLAN~\cite{zhuang2018parallel} & \bf 75.31 & 65.52 & 61.34 & 50.86 & 58.03\\
\hline
\hline
VC w/o reg & 70.78 & 65.10 & 56.82 & 51.30 & 60.95 \\
\hline
VC w/o $\alpha$ & 70.73 & 64.63 & 53.33 & 46.88 & 55.72 \\
\hline
VC  & 73.33 & \bf 67.44 & 58.40 & \bf 53.18 & \bf 62.30 \\
\hline
\end{tabular}
}
\vspace{-4mm}
\end{table}

\subsection{Evaluations of Supervised Grounding}
We compared our variational context (VC) method with state-of-the-art referring expression methods published in recent years, which can be categorized into: 1) generation-comprehension based such as MMI~\cite{mao2016generation}, Attr~\cite{liureferring}, Speaker~\cite{yu2016joint}, Listener~\cite{yu2016joint}, and SCRC~\cite{hu2016natural}; 2) localization based such as GroundR~\cite{rohrbach2016grounding}, NegBag~\cite{nagaraja2016modeling}, CMN~\cite{hu2016modeling}, MAttNet~\cite{yu2018mattnet}, PLAN~\cite{zhuang2018parallel}, A-ATT~\cite{deng2018visual}. Note that NegBag and CMN are MIL-based (multiple instance learning) models. In particular, we used the author-released code to obtain the results of CMN on RefCLEF, RefCOCO, and RefCOCO+. 

\noindent \textbf{Single comprehension module.} From the results of VGG-based models on RefCOCO, RefCOCO+, and RefCOCOg in Table~\ref{tab:1-2} and ~\ref{tab:1-3}, and that on RefCLEF in Table~\ref{tab:3}, we can see that VC achieves the state-of-the-art performance. We believe that the improvement is attributed to the variational Bayesian modeling of context. First, on all datasets, except for the most recent reinforcement learning based method~\cite{yu2016joint} or multiple attention mechanism based mathods~\cite{yu2018mattnet,zhuang2018parallel,deng2018visual} on the Test A split, VC outperforms all the other sentence generation-comprehension methods that do not model context. Second, compared to VC without the regularization term in Eq.~\eqref{eq:3} (VC w/o reg), VC can boost the performance by around 2\% on all datasets. This demonstrates the effectiveness of the KL divergence for the prevention of the overfitted context estimation. 

\begin{figure*}
	\centering
	\includegraphics[width=1\linewidth]{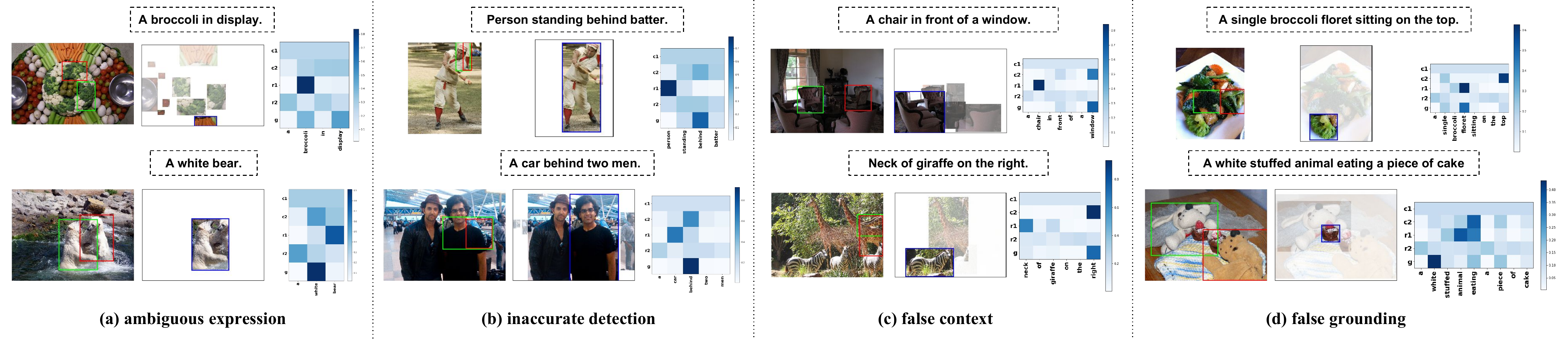}
	\caption{Common failure cases of our full model in supervised grounding on RefCOCOg. Each example shows grounding results, context estimation results, and cue-specific features from left to right. The denotations of the bounding box colors are as follows. Solid red: grounding referent; solid green: ground truth; solid blue: grounding context with highest probability.}
\vspace{-4mm}
\label{fig:7}
\end{figure*}

In particular, we further demonstrate the superiority of VC over the most recent MIL-based method CMN. As illustrated in Fig.~\ref{fig:4}, VC has better context comprehension in both of the language and image regions than CMN. For example, in the top two rows where VC is correct and CMN is wrong, for the grounding in the second column, CMN unnecessarily considers the ``girl'' as context but the expression only describes using ``elephant''; in the last column, CMN misses the key context ``frisbee''. Even in the failure cases where VC is wrong and CMN is correct, VC still localizes reasonable context. For example, in the fourth column, although CMN grounds the correct TV, it is based on incorrect context of other TVs; while VC can predict the correct context ``children''. In addition, we observed that most of the cases that CMN is better than VC involves multiple humans. This demonstrates that VC is better at grounding objects of different categories. VC is also effective in images with more objects. Fig.~\ref{fig:5} shows the performances of VC and CMN with various number of bounding boxes. We can observe that VC considerably outperforms CMN over all bounding boxes numbers. Recall that context is the key to distinguish objects of the same category. In particular, on the Test A sets of RefCOCO and RefCOCO+ where the grounding is only about people, \ie, the same object category, the gap between VC and CMN is becoming larger as the box number increases. This demonstrates that MIL is ineffective in modeling context, especially when the number of image regions is large.

\begin{table}
\centering
\caption{Performances (Acc\%) of supervised and unsupervised methods on RefCLEF. }
\label{tab:3}
\scalebox{0.88}{
\begin{tabular}{|c|c|c|c|}
\hline
                       & Sup. & Sup. (det) & Unsup. (det) \\ \hline
SCRC~\cite{hu2016natural}    &  72.74    &     17.93       &   ---           \\ \hline
GroundR~\cite{rohrbach2016grounding} &  ---    &     26.93       &   10.70           \\ \hline
CMN~\cite{hu2016modeling} & 81.52 & 28.33 &---  \\ \hline\hline
VC                     &   \textbf{82.43}   &     \textbf{31.13}     &   14.11   \\ \hline
VC w/o $\alpha$      &  79.60    &    27.40        &   \textbf{14.50}           \\ \hline
\end{tabular}
}
\vspace{-4mm}
\end{table}

\noindent \textbf{Cooperation with generation module.} Furthermore, we exploit the grounding performance of VC incorporating referring expression generation in the variational framework using Eq.~\eqref{eq:5-2}. As shown in Table~\ref{tab:4}, VC incorporating generation module using the policy gradient method REINFORCE (VC w/ Gen+PG) achieves the best performance except the Test B split of RefCOCO, which means that referring expression generation can help with comprehension in our framework. Note that we only use the single comprehension module during test time, which has the same structure with better-learned parameters compared to the single VC model. Note that VC w/ Gen slightly performs worse than VC on two splits. The possible reason comes from the difficulty of language understanding. The average length of referring expression in RefCOCO and RefCOCO+ is about 3.6, while the queries in RefCOCOg have an average length of around 8.4. This observation indicates that context estimation plays a more important role in grounding long descriptions than short phrases, since long description tends to include more complex context information.

\begin{table}[t]
\centering
\caption{Supervised grounding performances (Acc\%) of ablation study using generation module or better visual representation on MSCOCO ground-truth regions. $\dag$ and $\ddag$ indicates that this model uses res101 feature and attribute-based phrase-guided feature~\cite{yu2018mattnet}, respectively.}
\label{tab:4}
\scalebox{0.8}{
\begin{tabular}{|c|c|c|c|c|c|c|c|}
\hline 
&\multicolumn{2}{c|}{RefCOCO} & \multicolumn{2}{c|}{RefCOCO+} & \multicolumn{3}{c|}{RefCOCOg} \\
\hline
& Test A & Test B & Test A & Test B & Val* & Val & Test\\
\hline
MAttNet~\cite{yu2018mattnet}$^{\dag}$ & 81.58 & 83.34 & 66.59 & 65.08 & --- & 75.96 & 74.56\\
\hline
MAttNet~\cite{yu2018mattnet}$^{\dag\ddag}$ & 85.26 & 84.57 & 75.13 & 66.17 & --- & 78.10 & 78.12 \\
\hline
\hline
VC & 78.98 &  82.39 & 62.56 & 62.90 & 73.98 & 74.61 & 74.58\\
\hline
VC w/ Gen & 79.16 & 82.04 & 62.84 & 62.88 & 74.20 & 74.98 & 75.06\\
\hline
VC w/ Gen+PG &  79.30 & 82.04 & 63.22 & 63.12 & \bf 74.96 & 75.35 & 75.11 \\
\hline
VC w/ Gen+PG$^{\dag}$ & 80.40 & 83.51 & 67.52 & 66.46 & --- & 77.49 & 76.64\\
\hline
VC w/ Gen+PG$^{\dag\ddag}$ & \bf 86.26 & \bf 85.00 & \bf 76.48 & \bf 68.13 & --- &\bf 79.80 & \bf 79.96 \\
\hline
\end{tabular}
}
\vspace{-4mm}
\end{table}

\noindent \textbf{Better visual representation.} Recently, \cite{yu2018mattnet} use ResNet-FPN backbone for feature extraction instead of VGG. We also evaluate VC model using ResNet-based Faster-RCNN visual representation, which can further improve the grounding performance by at least 1\%, especially on RefCOCO+ dataset. In additional to ResNet feature, MAttNet also includes attention mechanism to obtain phrase-guided embedding in cooperation with object attribute prediction. For fair comparison, we just use their pre-trained model to extract attribute-based phrase-guided feature, and concatenate it with origin visual feature. As shown in Table \ref{tab:4}, VC w/ Gen+PG yields the state-of-the-art MAttNet by at least $1\%$ on RefCOCO+ and RefCOCOg datasets. It is worth noting that the relationship module in MAttNet assumes that only one object contributes to the context, which suffers from the ineffectiveness of MIL in modeling context.

\noindent \textbf{Qualitative results and failure cases studies.} The qualitative results of our full model on RefCOCOg dataset are shown in Fig.~\ref{fig:6}. As illustrated in Fig.~\ref{fig:6}, our full model estimates reasonable context and cue-specific feature. For example, for the referring expression ``a calf being bottle fed'', the word ``calf'' is the key to referent-cue feature. Since there are two calves in the image, our full model highlights the word ``fed'' in context-cue and generic-cue features. Meanwhile, our full model correctly focuses on the ``fed'' relation represented by the visual content of bottle and man in context estimation. Some common failure cases on RefCOCOg are illustrated in Figure \ref{fig:7}, classified into four groups: ambiguous expression, inaccurate detection, false context, and false grounding. Ambiguous expression means that the expression matches more than one candidate objects. Actually the grounding results in first row are correct since they also correspond to the referring expression. Inaccurate detection means that the referent is partially or even not detected due to the limitation of detector. Although our model successfully localizes the referent, only a part of target object has been detected. False context means that the context estimation result is incorrect. The failure estimated context confuses the grounding module to distinguish the ground-truth referent from other objects of the same category. False grounding means that the grounding module fails although the detection and estimated context are correct.

\begin{figure}[t]
	\centering
	\includegraphics[width=.9\linewidth]{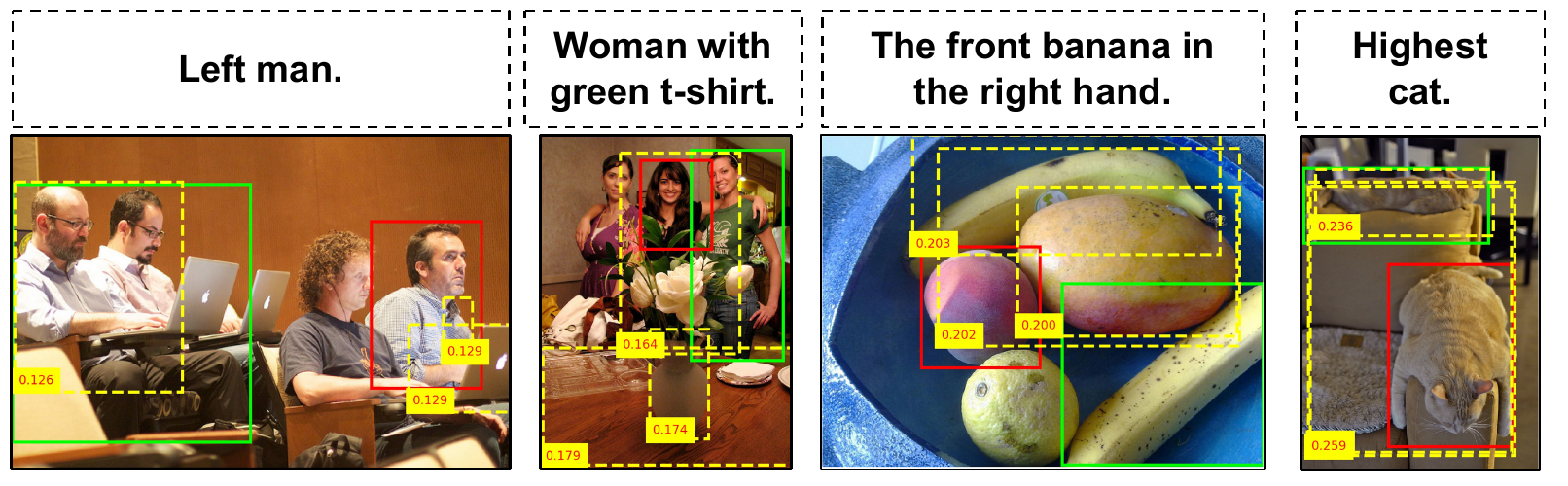}
	\caption{Common failure cases in unsupervised grounding with detected bounding boxes. From left to right: RefCOCO, RefCOCO+, and RefCOCOg. The failure is mainly to the challenging unsupervised relation modeling between referent and context.}
\vspace{-2mm}
\label{fig:8}
\end{figure}

\begin{table}[t]
\centering
\caption{Unsupervised grounding performances (Acc\%) of comparing methods using VGG features on RefCOCO, RefCOCO+, and RefCOCOg.}
\label{tab:5}
\scalebox{0.9}{
\begin{tabular}{|c|c|c|c|c|c|}
\hline 
Dataset & \multicolumn{2}{c|}{RefCOCO} & \multicolumn{2}{c|}{RefCOCO+} & RefCOCOg \\
\hline
Split & Test A & Test B & Test A & Test B & Val* \\
\hline
VC w/o reg & 13.59 & 21.65 & 18.79 & 24.14 & 25.14 \\
\hline
VC  & 17.34 & 20.98 & 23.24 & 24.91 & \bf 33.79 \\
\hline
VC w/o $\alpha$ & \bf 33.29 & \bf 30.13 & \bf 34.60 & \bf 31.58 & 30.26 \\
\hline
\hline 
Dataset & \multicolumn{2}{c|}{RefCOCO (det)} & \multicolumn{2}{c|}{RefCOCO+ (det)} & RefCOCOg (det) \\
\hline
Split & Test A & Test B & Test A & Test B & Val* \\
\hline
VC w/o reg & 17.14 & 22.30 & 19.74 & 24.05 & 28.14 \\
\hline
VC  & 20.91 & 21.77 & 25.79 & 25.54 & \bf 33.66 \\
\hline
VC w/o $\alpha$ & \bf 32.68 & \bf 27.22 & \bf 34.68 & \bf 28.10 & 29.65 \\
\hline
\end{tabular}
}
\vspace{-4mm}
\end{table}

\subsection{Evaluations of Unsupervised Grounding}

We follow the unsupervised setting in GroundR~\cite{rohrbach2016grounding}. To our best knowledge, it is the only work on unsupervised referring expression grounding.  Note that it is also known as ``weakly supervised'' detection~\cite{zhang2017ppr} as there is still image-level ground truth (\ie, the referring expression). Table~\ref{tab:3} reports the unsupervised results on the RefCLEF. We can see that VC outperforms the state-of-the-art GroundR, which is a generation-comprehension based method. This demonstrates that using context also helps unsupervised grounding. As there is no published unsupervised results on RefCOCO, RefCOCO+, and RefCOCOg, we only compared our baselines on them in Table~\ref{tab:5}. We can have the following three key observations which highlight the challenges of unsupervised grounding:

\begin{figure}[t]
	\centering
	\includegraphics[width=.9\linewidth]{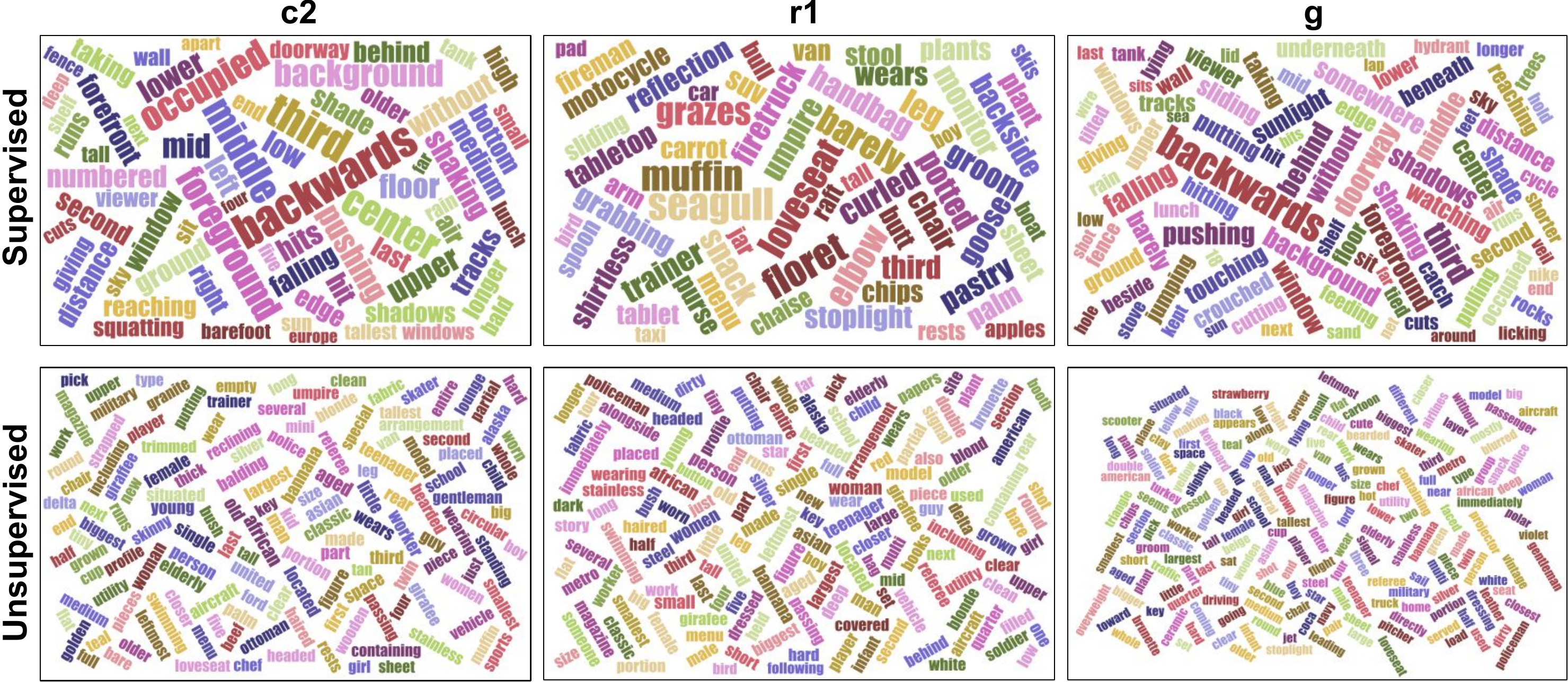}
	\caption{Word cloud visualizations of cue-specific word attention $\alpha$ in Eq.~\eqref{eq:10} of context-cue (c2), referent-cue (r1), and generic-cue (g) using supervised (top row) and unsupervised training (bottom row) on RefCOCOg. Without supervision, it is difficult to discover meaningful language compositions.}
\vspace{-3mm}
\label{fig:9}
\end{figure}

\textbf{Context Prior}. VC w/o reg is the baseline without the KL divergence as a context regularization in Eq.~\eqref{eq:3}. We can see that in most of the cases, VC considerably outperforms VC w/o reg by over 2\%, even over 5\% on RefCOCO+ (det) and RefCOCOg (det). Note that this improvement is significantly higher than that in supervised setting (\eg, $<3\%$ as reported in Table~\ref{tab:1-2}). The reason is that the context estimation in Eq.~\eqref{eq:4} would be easier to be stuck in image regions that are irrelevant to the expression in unsupervised setting, therefore, context prior is necessary. 

\textbf{Language Feature}. Except on RefCOCOg, we consistently observed the \emph{ineffectiveness} of the cue-specific language feature in unsupervised setting, \ie, VC w/o $\alpha$ outperforms VC in Table~\ref{tab:3} and~\ref{tab:5}. Here $\alpha$ represents the cue-specific word attention. This is contrary to the observation in the supervised setting as listed in Table~\ref{tab:1-2} and ~\ref{tab:1-3}, where VC w/o $\alpha$ is consistently lower than VC. Note that without the cue-specific word attention $\alpha$ in Eq.~\eqref{eq:10}, the language feature is merely the average value of the word embedding vectors in the expression. In this way, VC w/o $\alpha$ does not encode any structural language composition as illustrated in Fig.~\ref{fig:3}, thus, it is better for short expressions. However, when the expression is long in RefCOCOg, discarding the language structure still degrades the performance on RefCOCOg. 

\textbf{Unsupervised Relation Discovery}. Although we demonstrated that VC improves the unsupervised grounding by modeling context, we believe that there is still a large space for improving the quality of modeling the context. As the failure examples shown in Fig.~\ref{fig:8}, 1) many context estimations are still out of the scope of the expression, \eg, we may localize the ``cup'' and ``table'' as context even though the expression is ``woman with green t-shirt''; 2) we may mistake due to the wrong comprehension of the relations, \eg, ``right'' as ``left'', even if the objects belong to the same category, \eg, ``elephant''. For further investigation, Fig.~\ref{fig:9} visualizes the cue-specific word attentions in supervised and unsupervised settings. The almost identical word attentions in unsupervised setting reflect the fact that the relation modeling between referent and context is not as successful as in supervised setting. This inspires us to exploit stronger prior knowledge such as language structure~\cite{xiao2017cvpr} and spatial configurations~\cite{zhang2017ppr,wei2017object}.

\begin{table}[t]
\centering
\caption{Automatic metrics on referring expression generation. Note that \cite{yu2016joint} reports slightly higher accuracy using reranking mechanism. For fair comparison, we only report their performance without reranking.}
\label{tab:6}
\scalebox{0.85}{
\begin{tabular}{|c|c|c|c|c|c|}
\hline
\multirow{2}{*}{ } & \multicolumn{4}{c|}{ RefCOCO (Test A)} \\
\cline{2-5}
& BLEU-1 & BLEU-2 & METEOR & CIDEr \\
\hline
MMI~\cite{mao2016generation} & 0.478 & 0.295 & 0.175 & - \\
\hline
visdif~\cite{yu2016modeling} & 0.505 & 0.322 & 0.184 & - \\
\hline
Speaker~\cite{yu2016joint} & - & - & \bf 0.268 & 0.697 \\
\hline
Gen & 0.472 & 0.299 & 0.170 & 0.641 \\
\hline
VC w/ Gen & 0.548 & 0.361 & 0.188 & 0.707 \\
\hline
VC w/ Gen+PG & \bf 0.556 & \bf 0.368 & 0.194 & \bf 0.716 \\
\hline
\hline
\multirow{2}{*}{ } & \multicolumn{4}{c|}{ RefCOCO (Test B)} \\
\cline{2-5}
& BLEU-1 & BLEU-2 & METEOR & CIDEr \\
\hline
MMI~\cite{mao2016generation} & 0.547 & 0.341 & 0.228 & - \\
\hline
visdif~\cite{yu2016modeling} & 0.583 & 0.382 & 0.245 & - \\
\hline
Speaker~\cite{yu2016joint} & - & - & \bf 0.329 & 1.323 \\
\hline
Gen & 0.548 & 0.351 & 0.237 & 1.271 \\
\hline
VC w/ Gen & 0.628 & 0.424 & 0.245 & 1.356 \\
\hline
VC w/ Gen+PG & \bf 0.639 & \bf 0.430 & 0.252 & \bf 1.364 \\
\hline
\hline
\multirow{2}{*}{ } & \multicolumn{4}{c|}{ RefCOCO+ (Test A)} \\
\cline{2-5}
& BLEU-1 & BLEU-2 & METEOR & CIDEr \\
\hline
MMI~\cite{mao2016generation} & 0.370 & 0.203 & 0.136 & - \\
\hline
visdif~\cite{yu2016modeling} & 0.407 & 0.235 & 0.145 & - \\
\hline
Speaker~\cite{yu2016joint} & - & - & \bf 0.204 & 0.494 \\
\hline
Gen & 0.353 & 0.194 & 0.120 & 0.415 \\
\hline
VC w/ Gen & 0.426 & 0.229 & 0.142 & 0.518 \\
\hline
VC w/ Gen+PG & \bf 0.439 & \bf 0.235 & 0.151 & \bf 0.531 \\
\hline
\hline
\multirow{2}{*}{} & \multicolumn{4}{c|}{ RefCOCO+ (Test B)} \\
\cline{2-5}
& BLEU-1 & BLEU-2 & METEOR & CIDEr \\
\hline
MMI~\cite{mao2016generation} & 0.324 & 0.167 & 0.133 & - \\
\hline
visdif~\cite{yu2016modeling} & 0.339 & 0.177 & 0.145 & - \\
\hline
Speaker~\cite{yu2016joint} & - & - & \bf 0.202 & 0.709 \\
\hline
Gen & 0.364 & 0.172 & 0.128 & 0.659 \\
\hline
VC w/ Gen & 0.391 & 0.197 & 0.146 & 0.731\\
\hline
VC w/ Gen+PG & \bf 0.404 & \bf 0.209 & 0.154 & \bf 0.742 \\
\hline
\hline
\multirow{2}{*}{} & \multicolumn{4}{c|}{ RefCOCOg (Val*)} \\
\cline{2-5}
 & BLEU-1 & BLEU-2 & METEOR & CIDEr \\
\hline
MMI~\cite{mao2016generation} & 0.428 & 0.263 & 0.144 & - \\
\hline
visdif~\cite{yu2016modeling} & 0.442 & 0.277 & 0.151 & - \\
\hline
Speaker~\cite{yu2016joint} & - & - & \bf 0.154 & 0.592 \\
\hline
Gen & 0.398 & 0.233 & 0.108 & 0.504 \\
\hline
VC w/ Gen & 0.456 &	0.281 & 0.139 & 0.625 \\
\hline
VC w/ Gen+PG & \bf 0.467 & \bf 0.287 & 0.146 & \bf 0.630 \\
\hline
\end{tabular}
}
\vspace{-2mm}
\end{table}

\subsection{Evaluation of Generation}
For the generation task, we first evaluate our models using BLEU, METEOR and CIDEr, which are widely used evaluation metrics in generated description evaluation. The automatic evaluation results using above metrics are given in Table~\ref{tab:6}. Here, ``Gen'' represents the generation module trained separately without comprehension loss. From Table~\ref{tab:6}, we observe that the comprehension module helps to improve all the metrics significantly compared to the single generation module. This observation indicates that the estimated comprehension context can help to promote the performance of the generation module. In addition, our full model (VC w / Gen+PG) achieves the highest score on BLEU-1, BLEU-2 and CIDEr, while obtains slightly lower METEOR than Speaker.

\begin{figure*}
	\centering
	\includegraphics[width=.84\linewidth]{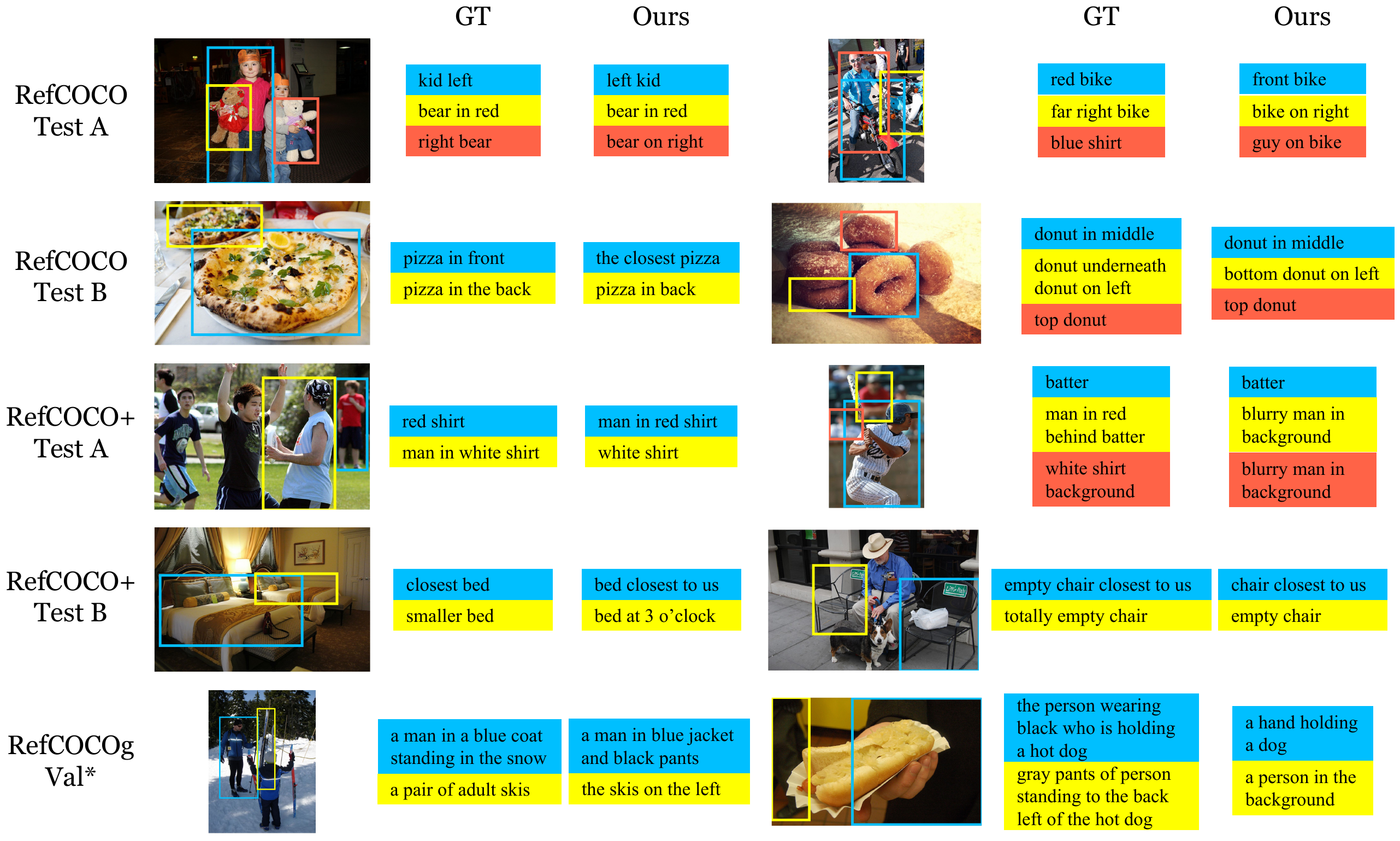}
	\caption{Example generation results using our full model (VC w / Gen+PG) on three datasets. The ground-truth/generated expression is linked with the described referent using the same color.}
\vspace{-4mm}
\label{fig:10}
\end{figure*}

Note that these metrics do not always reflect the ambiguity of generated description~\cite{yu2016modeling}, since there are multiple possible expressions which can distinguish one object from others. Thus, we follow \cite{yu2016modeling} and run a human evaluation on RefCOCO and RefCOCO+ to better evaluate the ambiguity of generated referring expression. Given the generated referring expression, the users were asked to click the referred object from the image, and the grounding accuracy was recorded for evaluation. The human evaluation results are shown in Table~\ref{tab:7}. Note that the comprehension module is only used for context estimation. The results show that generation module with estimated context has higher performance than vanilla generator, demonstrating that context modeling helps to generate unambiguous referring expression. In addition, the ResNet feature sightly improves the performance compared to the VGG feature. Fig.~\ref{fig:10} presents some example generation results on three datasets. Our full model is shown to be able to generate concise and unambiguous description with important context information, such as location (\eg, left, front), color (\eg, red, blue), and related objects (\eg, in red shirt, holding a dog). There are also cases that our model tends to fail. For the second example from the Test A split of RefCOCO+, our model succeeds to describe the audience in background using the clue ``blurry'' compared to batter in front, but fails to further distinguish the two audiences from color and location.
 
\begin{table}[t]
\centering
\caption{Human evaluation (Acc\%) on referring expression generation. Note that \cite{yu2016joint} reports slightly higher accuracy using reranking mechanism. For fair comparison, we only report their performance without reranking.}
\label{tab:7}
\scalebox{0.88}{
\begin{tabular}{|c|c|c|c|c|}
\hline 
&\multicolumn{2}{c|}{RefCOCO} & \multicolumn{2}{c|}{RefCOCO+} \\
\hline
& Test A & Test B & Test A & Test B \\
\hline
MMI \cite{mao2016generation} & 65.76 & 68.25 & 49.84 & 45.38 \\
\hline
Speaker \cite{yu2016joint} & 74.08 & 76.44 & 56.92 & 53.23 \\
\hline
\hline
Gen & 71.16 & 74.28 & 53.24 & 52.97\\
\hline
VC w/ Gen & 74.52 & 77.18 & 56.04 & 56.26\\
\hline
VC w/ Gen+PG & 74.39 & 77.56 & 56.35 & 56.48\\
\hline
VC w/ Gen+PG (resnet) & \bf 75.27 & \bf 78.62 & \bf 57.56 & \bf 57.83\\
\hline
\end{tabular}
}
\vspace{-4mm}
\end{table}

\section{Conclusions}
We focused on the task of grounding referring expressions in images and discussed that the key problem is how to model the complex context, which is not effectively resolved by the multiple instance learning framework used in prior works. Towards this challenge, we introduced the Variational Context framework, where the variational lower-bound can be interpreted by the reciprocity between the referent and context: given any of which can help to localize the other, and hence is expected to significantly reduce the context complexity in a principled way. The generation module is further included for semantic context modeling. The framework is implemented using cue-specific language-vision embedding network and policy gradient method that can be efficiently trained end-to-end. We validated the effectiveness of this reciprocity by promising supervised and unsupervised experiments on four benchmarks. We expect a future direction on one-stage visual grounding~\cite{chen2018real,deng2019you}, where the target object for the referring expression is directly localized without the region proposal generation stage for efficiency.

\ifCLASSOPTIONcaptionsoff
  \newpage
\fi

{\footnotesize
\bibliographystyle{ieee}
\bibliography{egbib}
}

\input{8-bio}

\end{document}

%% file: 8-bio.tex
\begin{IEEEbiography}[{\includegraphics[width=1in,height=1.25in,clip,keepaspectratio]{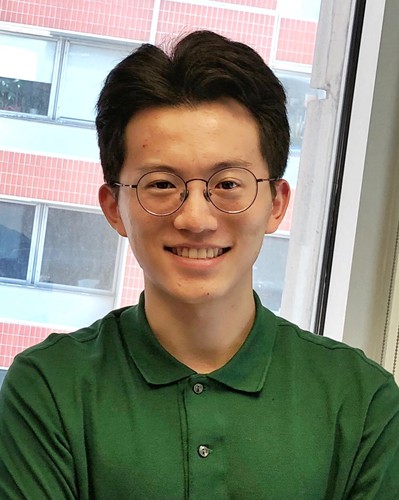}}]
{Yulei Niu} received the B.E. degree in computer science from the Renmin University of China, Beijing, China, in 2015, where he is currently pursuing the Ph.D. degree in computer science. From 2017 to 2018, he visited the Digital Video and Multimedia Laboratory, Columbia University, as a Visiting Ph.D. Student, under the supervision of Prof. Shih-Fu Chang. His research interests include computer vision, multimedia, and machine learning.
\end{IEEEbiography}

\begin{IEEEbiography}[{\includegraphics[width=1in,height=1.25in,clip,keepaspectratio]{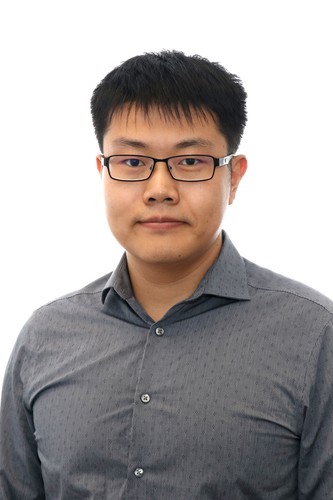}}]
{Hanwang Zhang} is currently an Assistant Professor at Nanyang Technological University, Singapore. He was a research scientist at the Department of Computer Science, Columbia University, USA. He has received the B.Eng (Hons.) degree in computer science from Zhejiang University, Hangzhou, China, in 2009, and the Ph.D. degree in computer science from the National University of Singapore in 2014. His research interest includes computer vision, multimedia, and social media. Dr. Zhang is the recipient of the Best Demo runner-up award in ACM MM 2012, the Best Student Paper award in ACM MM 2013, and the Best Paper Honorable Mention in ACM SIGIR 2016, and TOMM best paper award 2018. He is also the winner of Best Ph.D. Thesis Award of School of Computing, National University of Singapore, 2014.
\end{IEEEbiography}

\begin{IEEEbiography}[{\includegraphics[width=1in,height=1.25in,clip,keepaspectratio]{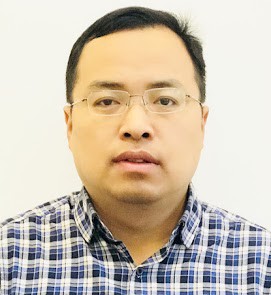}}]
{Zhiwu Lu} received the M.S. degree in applied mathematics from Peking University in 2005, and the Ph.D. degree in computer science from the City University of Hong Kong in 2011. He is currently an Associate Professor with the School of Information, Renmin University of China. He received the Best Paper Award at CGI 2014 and the IBM SUR Award 2015. His research interests include machine learning, pattern recognition, and computer vision.
\end{IEEEbiography}

\begin{IEEEbiography}[{\includegraphics[width=1in,height=1.25in,clip,keepaspectratio]{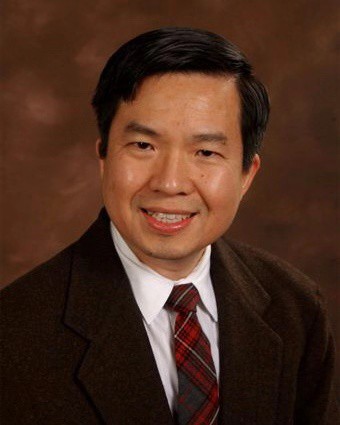}}]
{Shih-Fu Chang} is the Senior Executive Vice Dean of engineering with Columbia University and a Professor in electrical engineering and computer science. His research interests include computer vision, machine learning, and multimedia information retrieval, with the goal to turn unstructured multimedia data into searchable information. He is a fellow of the American Association for the Advancement of Science and the ACM.
\end{IEEEbiography}